\documentclass[10pt,twocolumn,letterpaper]{article}

\usepackage{cvpr}
\usepackage{times}
\usepackage{epsfig}
\usepackage{graphicx}
\usepackage{amsmath}
\usepackage{amssymb}
\usepackage{bbm}
\usepackage{multirow}
\usepackage{flushend}
\usepackage{booktabs}
\usepackage{comment}
\usepackage{algorithm}
\usepackage{algpseudocode}
\usepackage[sort,nocompress]{cite}

\usepackage{pifont}
\newcommand{\cmark}{\text{\ding{51}}}
\newcommand{\xmark}{\text{\ding{55}}}

\usepackage[pagebackref=true,breaklinks=true,letterpaper=true,colorlinks,bookmarks=false]{hyperref}

\cvprfinalcopy 

\begin{document}

\title{Activity Graph Transformer for Temporal Action Localization}

\author{Megha Nawhal$^{1}$, Greg Mori$^{1,2}$\\
$^{1}$ Simon Fraser University, Burnaby, Canada\\
$^{2}$ Borealis AI, Vancouver, Canada\\
}

\maketitle
\thispagestyle{empty}


\begin{abstract}
We introduce Activity Graph Transformer, an end-to-end learnable model for temporal action localization, that receives a video as input and directly predicts a set of action instances that appear in the video.
Detecting and localizing action instances in untrimmed videos requires reasoning over multiple action instances in a video. The dominant paradigms in the literature process videos temporally to either propose action regions or directly produce frame-level detections. However, sequential processing of videos is problematic when the action instances have non-sequential dependencies and/or non-linear temporal ordering, such as overlapping action instances or re-occurrence of action instances over the course of the video. In this work, we capture this non-linear temporal structure by reasoning over the videos as non-sequential entities in the form of graphs.  We evaluate our model on challenging datasets: THUMOS14, Charades, and EPIC-Kitchens-100. Our results show that our proposed model outperforms the state-of-the-art by a considerable margin.

\end{abstract}

\section{Introduction}

\begin{figure}[t]
    \centering
    \includegraphics[width=0.49\textwidth]{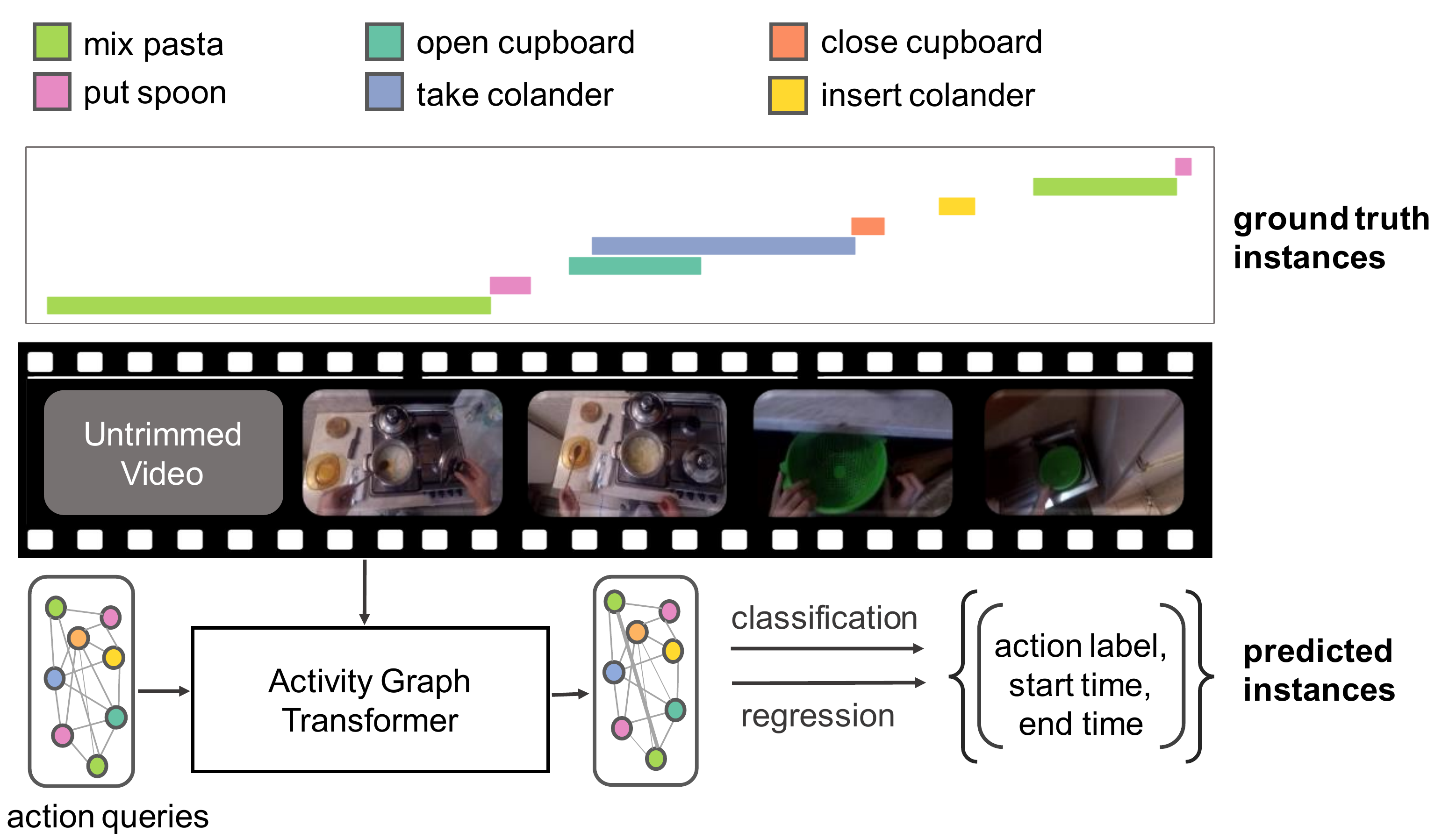}
    \caption{\textbf{Main Idea.} Given an untrimmed human activity video, we directly predict the set of action instances (label, start time, end time) that appear in the video. We observe that human activity videos contain non-sequential dependencies (illustrated by the ground truth instances as colored bars). In this work, we propose Activity Graph Transformer that captures this non-sequential structure by reasoning over such videos as graphs. Overall, the network receives a video and directly infers a set of action instances. The network achieves this by transforming a set of graph-structured abstract queries into contextual embeddings which are then used to provide predictions of action instances. It is trained end-to-end using classification and regression losses. }
    \label{fig:teaser}
\end{figure}

Visual understanding of human activities in untrimmed videos involves reasoning over multiple action instances with varying temporal extents. This problem has been formally studied in the setup of temporal action localization,  \ie, given a human activity video, the goal is to predict a set of action labels and their corresponding start and end timestamps indicating their occurrence in the video. Reasoning over untrimmed human activity videos for action localization is particularly challenging due to the idiosyncrasies of the videos 
such as: (1) overlap - the action instances may have overlaps in their temporal extents indicating non-sequential temporal ordering of the instances; (2) non-sequential dependencies - some action instances may have temporal dependencies but are separated by other unrelated action instances and/or durations of no action; and (3) re-occurrence - instances belonging to same category may appear more than once over the course of the video. 
In this work, we propose a novel end-to-end learnable model for temporal action localization that receives a video as an input and directly predicts the set of action instances that appear in the video. 

Existing approaches for the task of temporal action localization predominantly fall into two paradigms. First is the local-then-global paradigm where the video-level predictions are obtained by postprocessing of local (\ie frame-level or snippet-level) predictions using sequence modeling techniques such as recurrent neural networks, temporal convolutions and temporal pooling~\cite{richard2016temporal,yuan2017temporal,dave2017predictive, singh2016multi, ma2016learning, yeung2016end,lea2017temporal,piergiovanni2018learning,karpathy2014large,piergiovanni2018learning}. Second is the proposal-then-classification paradigm which involves generation of a sparse set of class agnostic segment proposals from the overall video followed by classification of the action categories for each proposal using either two-stage learning~\cite{caba2016fast,buch2017sst,heilbron2017scc,escorcia2016daps,shou2016temporal,shou2017cdc,zhao2017temporal,zhao2020bottom} or end-to-end learning~\cite{dai2017temporal,gao2017turn,chao2018rethinking,xu2017r,zeng2019graph}. 

The local-then-global paradigm does not utilize the overall temporal context provided by the activity in the video as the local predictions are solely based on visual information confined to the frame or the snippet. For instance, consider the example in Figure~\ref{fig:teaser}, these approaches would miss out on important relevant information provided by \textit{`mix pasta'} in predicting \textit{`put spoon'} or may produce imprecise predictions when the temporal extents of instances \textit{`take colander'} or \textit{`open cupboard'} overlap. 

Alternatively, the proposal-then-classification paradigm generates a subset of proposals by processing the video as a sequence. As a result, these approaches suffer from limited receptive field for incorporating temporal information, and do not capture non-sequential temporal dependencies effectively. This problem is further aggravated in the case of overlapping action instances.
For instance, in the example in Figure~\ref{fig:teaser}, \textit{`open cupboard'} and \textit{`close cupboard'} share information but are separated by other, potentially overlapping, action instances such as \textit{`take colander'}. Due to such ordering, when generating proposals corresponding to \textit{`close cupboard'}, these approaches are unlikely to capture the dependency with the visual information pertaining to \textit{`open cupboard'}.
Furthermore, these approaches use heuristics to perform non-maximal suppression of proposals  that might result in imprecise localization outcomes when the action instances vary widely in their temporal extents. 

As such, both these types of approaches process videos sequentially to either generate direct local predictions or action proposals and are problematic when action instances reoccur, overlap, or have non-sequential dependencies.
These observations suggest that although a video has a linear ordering of frames, the reasoning over the video need not be sequential. We argue that modeling the non-linear temporal structure is a key requirement for effective reasoning over untrimmed human activity videos. In this work, we seek a temporal action localization model that: (1) captures the temporal structure in complex human activity videos, (2) does not rely on heuristics or postprocessing of the predictions, and (3) is trained end-to-end.

Towards this goal, we formulate temporal action localization as a direct set prediction task. 
We propose a novel temporal action localization model, \textit{Activity Graph Transformer} (AGT), an end-to-end learnable model that receives a video as input and predicts the set of action instances that appear in the video. In order to capture the non-linear temporal structure in videos, we reason over videos as non-sequential entities, specifically, learnable graph structures. 
Particularly, we map the input video to graph-structured embeddings using an encoder-decoder transformer architecture that operates using graph attention. A final feed forward network then uses these embeddings to directly predict the action instances.
Thus, we propose a streamlined end-to-end training process that does not require any heuristics.

To summarize, our contributions are as follows: (1) we propose an encoder-decoder transformer based model Activity Graph Transformer that reasons over videos as graphs and can be trained end-to-end, and (2) we achieve state-of-the-art performance on the task of temporal action localization on challenging human activity datasets, namely, THUMOS14~\cite{THUMOS14}, Charades~\cite{sigurdsson2016hollywood}, and EPIC-Kitchens100~\cite{Damen2020RESCALING}. 


\section{Related Work}
\noindent
In this section, we discuss the prior work relevant to temporal action localization and graph based modeling in videos.

\vspace{0.05in}
\noindent
\textbf{Temporal Action Localization.} Early methods for temporal action localization use temporal sliding windows and design hand-crafted features to classify action within each window~\cite{yuan2016temporal,gaidon2013temporal,jain2014action,tang2013combining,oneata2013action}. However, these approaches are computationally inefficient as they apply classifiers on windows of all possible sizes and locations in the entire video. 

With the advances in convolutional neural networks, recent approaches fall into two dominant paradigms: (1) local-then-global, (2) proposal-then-classification. The methods following the local-then-global paradigm rely on obtaining temporal boundaries of actions based on local (\ie frame-level or snippet-level) predictions and perform video-level reasoning using temporal modeling techniques such as explicit modeling of action durations or transitions~\cite{richard2016temporal,yuan2017temporal}, recurrent neural networks~\cite{dave2017predictive, singh2016multi, ma2016learning, yeung2016end}, temporal pooling~\cite{karpathy2014large}, temporal convolutions~\cite{lea2017temporal,piergiovanni2018learning}, and temporal attention~\cite{piergiovanni2018learning}. However, these approaches does not utilize the overall temporal context of the videos as local predictions are computed using only the frame/snippet information.

The methods based on proposal-then-classification paradigm formulate temporal action localization as the mirror problem of object detection in the temporal domain. Inspired by the progress in object detection~\cite{girshick2015fast} techniques, some methods employ a two-stage training framework ~\cite{caba2016fast,buch2017sst,heilbron2017scc,escorcia2016daps,shou2016temporal,shou2017cdc,zhao2017temporal,zhao2020bottom} -- they generate a set of class-agnostic segment proposals in the first stage and predict an action label for each proposal in the second stage. Most recent methods in this direction focus on improving the proposal generation stage ~\cite{caba2016fast,buch2017sst,heilbron2017scc,escorcia2016daps,lin2019bmn,lin2018bsn,zhao2020bottom,bai2020boundary}, while a few propose a more accurate classification stage~\cite{shou2017cdc,zhao2017temporal}. 

Recently, some end-to-end trainable architectures have also been proposed~\cite{dai2017temporal,gao2017turn,chao2018rethinking,xu2017r,zeng2019graph,xu2020g}. However, these methods also process the video as a sequence and, thus, have limited receptive field for capturing temporal information. They do not capture non-sequential temporal dependencies in action instances. Moreover, these approaches use heuristics during training (\eg intersection-over-union thresholds) to perform non-maximal suppression in the set of proposals. This might lead to poor localization performance when the action instances vary widely in their temporal extents as it might skip some highly overlapping proposals. To address these problems in object detection, ~\cite{carion2020end} propose a transformer based end-to-end learnable architecture that implicitly learns the non-max suppression and perform object detection using proposals as abstract encodings. 

In contrast to the above approaches, we formulate temporal action localization as a direct set prediction task. We propose to reason over untrimmed videos as non-sequential entities (\ie graphs) as opposed to existing methods that perform sequential reasoning. Our approach is inspired by ~\cite{carion2020end} in that we propose an end-to-end learnable transformer based model for direct set prediction. But unlike ~\cite{carion2020end}, the transformer model in our approach operates graphs. 

Additionally, there are other realms of work on temporal action localization in weakly supervised setting~\cite{Shou_2018_ECCV,wang2017untrimmednets,jain2020actionbytes} and spatio-temporal action localization~\cite{singh2017online,kalogeiton2017action,gkioxari2015finding,girdhar2019video}. These are beyond the scope of this paper.

\vspace{0.05in}
\noindent
\textbf{Action Recognition.} Action recognition methods operate on short video clips that are trimmed such that a single action instance spans the video duration and, hence, are not suitable for untrimmed videos containing multiple actions. Nonetheless, models pretrained for the task of action recognition provide effective feature representations for tasks related to untrimmed videos. A wide variety of action recognition approaches have been proposed ranging from earlier methods based on hand-crafted features~\cite{laptev2005space,dalal2006human,wang2013action} to convolutional models such as I3D ~\cite{simonyan2014two}, 3D-CNN ~\cite{tran2015learning} through to advanced temporal modeling~\cite{wang2015action,wang2016temporal,zhang2016real} and graph modeling~\cite{wang2018videos,jain2016structural} techniques. In this paper, we use I3D ~\cite{simonyan2014two} pretrained on the Kinetics dataset~\cite{carreira2017quo} for feature extraction.

\vspace{0.05in}
\noindent
\textbf{Graph-based Modeling for Videos.} 
The advances in graph convolutional networks (GCNs) ~\cite{kipf2017semi} have inspired several recent approaches for video based tasks ~\cite{pan2020spatio,wang2018videos,nagarajan2020ego}. Most of the graph based approaches for videos represent either the input space (\ie videos or derived visual information) as graphs~\cite{pan2020spatio,wang2018videos,nagarajan2020ego,hussein2019videograph} or the output space (\ie labels) as graphs~\cite{tsai2019GSTEG}. In contrast, we design our model based on the insight that both the input space (\ie features derived from videos) and the output space (\ie labels and timestamps for the action) are graph-structured for the task of temporal action localization. Specifically, we propose an encoder-decoder transformer architecture to learn the mapping between the input and output space. Furthermore, GCNs require the information pertaining to the nodes and edges a priori. In contrast, we learn the graph structure (\ie both nodes and edges) from the data itself using self-attention. 

\section{Proposed Approach}
\begin{figure*}[h]
    \centering
    \includegraphics[width=\textwidth]{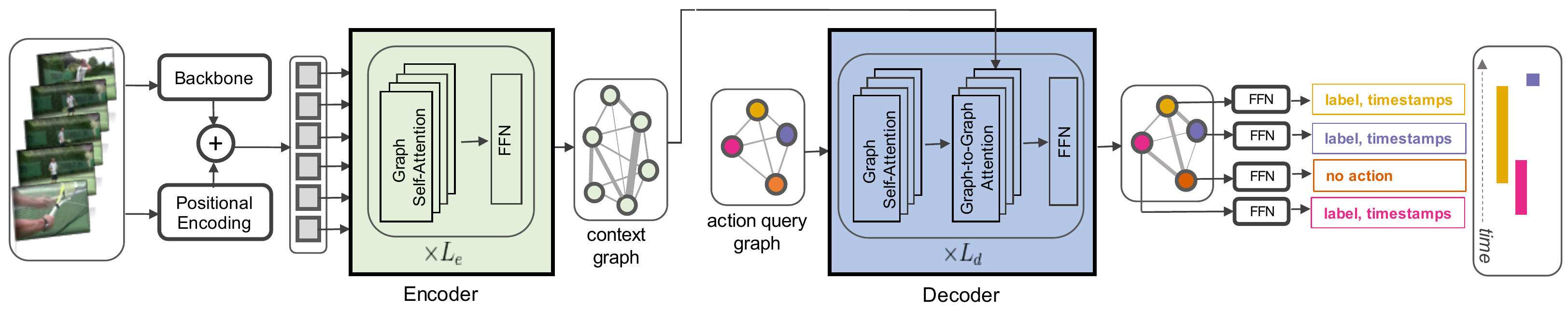}
    \caption{\textbf{Model Overview.} Activity Graph Transformer (AGT) receives a video as input and directly predicts a set of action instances that appear in the video.
    The input video is fed into a backbone network to obtain a compact representation (Section ~\ref{subsec:backbone}). Then, the encoder network (Section ~\ref{subsec:encoder}) receives the compact video-level representation from the backbone network and encodes it to a latent graph representation \textit{context graph}. The decoder network (Section ~\ref{subsec:decoder}) receives the context graph along with graph-structured abstract query encodings \textit{action query graph}. The decoder transforms the action query graph to a graph-structured set of embeddings.
    Each node embedding of the decoder output is fed into a prediction head (Section ~\ref{subsec:ffn}). The network is trained end-to-end (Section ~\ref{subsec:loss}) using classification and regression losses for the action labels and timestamps of the action instances respectively.}
    \label{fig:model}
\end{figure*}
In this section, we present the problem formulation and provide a detailed description of our proposed approach.

\vspace{0.05in}
\noindent
\textbf{Problem Formulation.} 
The task of temporal action localization involves prediction of the category labels as well as start and end timestamps of the actions that occur in a given video. In this work, we formulate this task as a direct set prediction problem wherein each element in the predicted set denotes an action instance in a video. 
Specifically, given a video $V$, the goal is to predict a set $\mathcal{A}$ where the $i$-th element $a^{(i)} = (c^{(i)}, t^{(i)}_s, t^{(i)}_e)$ denotes an action instance in the video depicting action category $c^{(i)}$ that starts at time $0 \leq t^{(i)}_s \leq T$, ends at time $0 \leq t^{(i)}_e \leq T$, for $i \in \{1, 2, \ldots, |\mathcal{A}|\}$. Here, $|\mathcal{A}|$ is the number of action instances present in the video and $T$ is the duration of the video. Thus, $|\mathcal{A}|$ and $T$ vary based on the input video.

Towards this goal, we propose \textit{Activity Graph Transformer} (AGT), an end-to-end learnable model that receives a video as input and directly infers the set of action instances (\textit{label, start time, end time}) in the video. 
Our approach consists of: (1) a network that predicts a set of action instances in a single forward pass; and (2) a loss function to train the network by obtaining a unique alignment between the predicted and ground truth action instances. 
We contend that effective reasoning over untrimmed human activity videos requires modeling the non-linear temporal structure in the videos. In our approach, we seek to capture this structure by employing graphs. 
Specifically, we propose a novel encoder-decoder transformer network
that leverages graph based self-attention to reason over the videos.
We describe the details of our approach below. 


\subsection{Activity Graph Transformer}
As shown in Figure~\ref{fig:model}, Activity Graph Transformer (AGT) consists of three components: (1) backbone network to obtain a compact representation of the input video; (2) transformer network consisting of an encoder network and a decoder network that operates over graphs; and (3) prediction heads for the final prediction of action instances of the form (label, start time, end time). The encoder network receives the compact video-level representation from the backbone network and encodes it to a latent graph representation, referred to as \textit{context graph}. The decoder network receives graph-structured abstract query encodings (referred to as \textit{action query graph}) as input along with the context graph. The decoder uses the context graph to transform the action query graph to a graph-structured set of embeddings.
Each node embedding of this decoder output is fed into a feed forward network to obtain predictions of action instances. The whole AGT network is trained end-to-end using a combination of classification and regression losses for the action labels and timestamps respectively. Refer to Algorithm~\ref{algo} for an overview of one training iteration of AGT. We provide detailed description of the components below. 

\subsubsection{Backbone} \label{subsec:backbone}
To obtain a compact representation for the input video $V$ containing $T$ frames, any 3D convolutional network can be used to extract the features. In our implementation, we chunk the videos into short overlapping segments of 8 frames and use an I3D model pretrained on the Kinetics\cite{carreira2017quo} dataset to extract features of dimension $C$ ($=2048$) from the segments, resulting in video-level feature $\mathbf{v} = [\mathbf{v}^{(1)},\mathbf{v}^{(2)} \ldots \mathbf{v}^{(N_v)}]$ where $N_v$ is the number of chunks used in the feature extraction. 

\subsubsection{Transformer Encoder}\label{subsec:encoder}
The backbone simply provides a sequence of local features and does not incorporate the overall context of the video or the temporal structure in the video. Therefore, we use an encoder network that receives the video-level feature as input and encodes this video representation to a graph (referred to as the \textit{context graph}). Intuitively, the encoder is designed to model the interactions among the local features using self-attention modules. 

The context graph is initialized with video-level feature $\mathbf{v}^{(i)}$ (of dimension $C=2048$) as the $i$-th node for $i \in \{1,2,\ldots,N_v\}$. Usually, transformer networks use fixed positional encoding~\cite{parmar2018image} to provide position information of each element in the input sequence. In contrast, in our setting, we contend that the video features have a non-linear temporal structure.
Thus, we provide the positional information using learnable positional encodings $\mathbf{p}_{v}$ as additional information to the video feature $\mathbf{v}$. The positional encoding $\mathbf{p}_{v}^{(i)}$ corresponds to the $i$-th node in the graph and is of the same dimension as the node.
Next, the graph nodes are from the same video and hence, they are related to each other. However, their connection information (edges) is not known a priori. Thus, we model the interactions among these nodes as learnable edge weights. This is enabled by the graph self-attention module (described below). 

We design the transformer encoder network $\mathbf{E}$ as a sequence of $L_e$ blocks, wherein, an  encoder block $\mathbf{E}_\ell$ for $\ell \in \{1,2,\ldots,L_{e}\}$ consists of a graph self-attention module  followed by 
a feed forward network. 
The output of the encoder network is the context graph $\mathbf{h}_{L_e} = [\mathbf{h}^{(1)}_{L_e}, \mathbf{h}^{(2)}_{L_e} \ldots \mathbf{h}^{(N_v)}_{L_e}]$ where $\mathbf{h}^{(i)}_{L_e}$ is the $i$-th node and is of dimension $d$ (same for each block). The output of the $\ell$-th encoder block $\mathbf{h}_{\ell}$ 
and the final output of the encoder $\mathbf{h}_{L_e}$ are defined as:
\begin{equation}
\label{eq:enc}
\begin{split}
    \mathbf{h}_{0} &= \mathbf{v}\\
    \mathbf{h}_{\ell} &= \mathbf{E}_{\ell}(\mathbf{h}_{\ell-1},\mathbf{p}_{v}) \\
    \mathbf{h}_{L_e} &= \mathbf{E}_{L_{e}} \circ \cdots \circ \mathbf{E}_{1} (\mathbf{v},\mathbf{p}_{v}).
\end{split}
\end{equation}

\vspace{0.05in}
\noindent
\textbf{Graph Self-Attention.} This module aims to model interactions among graph structured variables along with learnable edge weights. Here, we describe the graph self-attention module in the context of the $(\ell+1)$-th encoder block $\mathbf{E}_{\ell+1}$. For simplicity of notation, let $\mathbf{x}$ be the output of the $\ell$-th block of the encoder, \ie, $\mathbf{x} = \mathbf{E}_{\ell}(\mathbf{h}_{\ell-1},\mathbf{p}_{v})$. $\mathbf{x}$ is a graph contains $N_v$ nodes $\mathbf{x}^{(1)}, \mathbf{x}^{(2)}, \ldots, \mathbf{x}^{(N_v)}$ which are connected using learnable edge weights. The graph self-attention module first performs graph message passing (as described in ~\cite{velivckovic2017graph}) to produce the output $\mathbf{x}'$, with the $i$-th node of the output defined as
\begin{equation}
    \mathbf{x'}^{(i)} = \mathbf{x}^{(i)} + \Big| \Big|_{k=1} ^{K} \sigma \Big(\sum_{j \in \mathcal{N}_i} \alpha_{ij}^{k} \mathbf{W}_{g}^{k} \mathbf{x}^{(j)}\Big),
\end{equation}
where $\Big| \Big|$ represents concatenation operator, $K$ is the number of parallel heads in the self-attention module, $\sigma$ is a non-linear function (leaky ReLU in our case), $\mathcal{N}_i$ represents the set of neighbours of the $i$-th node, $\mathbf{W}_{g}^{k}$ is the learnable transformation weight matrix. $\alpha_{ij}^{k}$ are the self attention coefficients computed by the $k$-th attention head described as:
\begin{equation} \label{eq:graph_sa}
\alpha_{ij}^{k} = \frac{\exp (f(\mathbf{w}_{a,k}^{T} [\mathbf{W}_{g}^{k}\mathbf{x}^{(i)} || \mathbf{W}_{g}^{k}\mathbf{x}^{(j)}]))}{\sum_{m \in \mathcal{N}_i}\exp (f(\mathbf{w}_{a,k}^{T} [\mathbf{W}_{g}^{k}\mathbf{x}^{(i)} || \mathbf{W}_{g}^{k}\mathbf{x}^{(m)}]))},    
\end{equation}
where $\cdot^{T}$ represents a transpose operator, 
$f$ is a non-linear activation (leaky ReLU in our case) and $\mathbf{w}_{a,k}$ is the attention coefficients. $\alpha_{ij}^{k}$ is the attention weight and denotes the strength of the interaction between $i$-th and $j$-th node of the input graph of the module.
Subsequent to the message passing step, we apply batch normalization and a linear layer. This is then followed by a standard multi-head self-attention layer (same as in~\cite{vaswani2017attention}). 
Overall, the graph self-attention module models interactions between the nodes, \ie, local features derived from the video. 

\begin{algorithm}[t]
\caption{A training iteration of AGT model}
\label{algo}
\begin{algorithmic}[1]
\renewcommand{\algorithmicensure}{\textbf{Inputs:}}
\Ensure video $V$ containing $T$ frames; number of action query encodings $N_o$; ground truth action instances $\mathcal{A} = \{a^{(i)}\}_{i=1}^{|\mathcal{A}|}$
\renewcommand{\algorithmicensure}{\textbf{Initializations:}}
\Ensure Backbone initialized to I3D model~\cite{carreira2017quo} pretrained on Kinetics dataset; initialize action query graph $\mathbf{q}$ with $N_o$ random encodings

\State compute features using \textbf{backbone}\;
\State compute context graph using \textbf{encoder} (see Eq.~\ref{eq:enc})\;
\State compute output embeddings using \textbf{decoder} (see Eq.~\ref{eq:dec})\;
\For{$i \in 1,2,\ldots,N_o$}
    \State predict action instance $\tilde a^{(i)}$ using prediction head \;
    \EndFor
\State compute optimal matching $\hat \phi$ between $\{a^{(i)}\}_{i=1}^{|\mathcal{A}|}$ and $\{\tilde a^{(i)}\}_{i=1}^{N_o}$ using \textbf{matcher} \;
\State compute final loss $\mathcal{L}_{H}$ between $\{a^{(i)}\}_{i=1}^{|\mathcal{A}|}$ and $\{\tilde a^{(\hat\phi(i))}\}_{i=1}^{|\mathcal{A}|}$ using Eq.~\ref{eq:loss}\;
\State backpropagate $\mathcal{L}_{H}$
\end{algorithmic}
\end{algorithm}

\subsubsection{Transformer Decoder}\label{subsec:decoder}
Based on the observation that the action instances in the video have a non-linear temporal structure, we design the decoder to learn a graph-structured set of embeddings which would subsequently be used for predicting the action instances.
Intuitively, the output graph provided by the decoder
serves as the latent representation for the set of action instances depicted in the video. 

The inputs of the transformer decoder are: (1) a graph-structured abstract query encodings, referred to as action query graph $\mathbf{q}$, containing $N_o$ nodes wherein each node is a learnable positional encoding of dimension $d$ (same as the dimension used in the encoder); and (2) the context graph $\mathbf{h}_{L_e}$ containing $N_v$ nodes (obtained from the encoder). 
We assume that the number of nodes in the action query graph $N_o$ is fixed and is sufficiently larger than the maximum number of action instances per video in the dataset. This idea of using representations of prediction entities as positional query encodings is inspired from ~\cite{carion2020end}. However, unlike the independent queries in ~\cite{carion2020end}, we use graph-structured encodings for the decoder.
To learn the interactions among the graph-structured query embeddings, we use graph self-attention modules (same module as used in transformer encoder). 
Additionally, we use graph-to-graph attention module (described below) to learn interactions between the context graph, \ie, the latent representation of the input video, and the graph-structured query embeddings, \ie, the latent representations of the action queries.

The overall decoder network $\mathbf{D}$ consists of $L_d$ blocks, wherein, a decoder block $\mathbf{D}_{\ell'}$ for $\ell' \in \{1,2,\ldots,L_{d}\}$ consists of a graph self-attention module followed by a graph-to-graph attention module, and then a feed forward network. The decoder block $\mathbf{D}_{\ell'}$ has an output $\mathbf{y}_{\ell'}$ and the final output of the decoder $\mathbf{y}_{L_d} = [\mathbf{y}^{(1)}_{L_d}, \mathbf{y}^{(2)}_{L_d}, \ldots, \mathbf{y}^{(N_o)}_{L_d}]$. They are defined as:
\begin{equation}
\label{eq:dec}
\begin{split}
    \mathbf{y}_{0} &= \mathbf{q}\\
    \mathbf{y}_{\ell'} &= \mathbf{D}_{\ell'}(\mathbf{y}_{\ell'-1},\mathbf{h}_{L_e}) \\
    \mathbf{y}_{L_d} &= \mathbf{D}_{L_{d}} \circ \cdots \circ \mathbf{D}_{1} (\mathbf{q},\mathbf{h}_{L_{e}})
\end{split}
\end{equation}

\vspace{0.05in}
\noindent
\textbf{Graph-to-Graph Attention.} The graph-to-graph attention module aims to learn the interactions between two different graphs referred to as a source graph and a target graph. 
Here, we describe this module in the context of the decoder block $\mathbf{D}_{\ell'+1}$.  
The input to this block is the output $\mathbf{y}_{\ell'}$ of the previous decoder block $\mathbf{D}_{\ell'}$.
This is fed to the graph self-attention module in the  block $\mathbf{D}_{\ell'+1}$, and the output is used as the target graph for the graph-to-graph attention module. 
The source graph for this module (in any decoder block) is the context graph $\mathbf{h}_{L_e}$.
For simplicity of notation, 
let $\mathbf{x}_{s}$ denote the source graph (\ie $\mathbf{h}_{L_e}$) and $\mathbf{x}_{t}$ denote the target graph. 
Here, the source and target graphs may contain different number of nodes. In our case, $\mathbf{x}_{s}$ contains $N_v$ nodes and $\mathbf{x}_{t}$ contains $N_o$ nodes.
The graph-to-graph attention module first performs message passing from source graph to target graph
to provide an output $\mathbf{x'}_{t}$, with the $i$-th node defined as
\begin{equation}
    \mathbf{x'}_{t}^{(i)} = \mathbf{x}_{t}^{(i)} + \Big| \Big|_{k=1} ^{K} \sigma \Big(\sum_{j \in \mathcal{N}_i} \beta_{ij}^{k} \mathbf{W}_{s}^k \mathbf{x}_{s}^{(j)}\Big),
\end{equation}
where $\mathbf{W}_{s}^{k}$ is the learnable transformation weight matrix for the source graph.
Other symbols denote the same entities as in graph self-attention (Section~\ref{subsec:encoder}).
$\beta_{ij}^{k}$ is the attention coefficient for $k$-th attention head between $i$-th node of the source graph and $j$-th node of the target graph computed as:
\begin{equation}
    \beta_{ij}^k = \frac{\exp (f(\mathbf{w}^{st}{_{a,k}^{T}} [\mathbf{W}_{s}^{k}\mathbf{x}_{s}^{(i)} || \mathbf{W}_{t}^{k}\mathbf{x}_{t}^{(j)}]))}{\sum_{m \in \mathcal{N}_i}\exp (f(\mathbf{w}^{st}{_{a,k}^T} [\mathbf{W}_{s}^{k}\mathbf{x}_{s}^{(i)} || \mathbf{W}_{t}^{k}\mathbf{x}_{t}^{(m)}]))}
\end{equation}
where $\mathbf{w}^{st}_{a,k}$ is the graph-to-graph attention coefficients, and
$\mathbf{W}_{s}^{k}$ and
$\mathbf{W}_{t}^{k}$ are the learnable transformation weight matrices for source and target graphs respectively. 
Other symbols denote the same entities as in graph self-attention.
Similar to the transformer encoder, subsequent to the message passing step, we apply batch normalization and a linear layer. This is then followed by a standard multi-head self-attention layer (same as in~\cite{vaswani2017attention}). 
Overall, the graph-to-graph attention module models the interactions between the latent representations of the input video and the action queries.

\subsubsection{Prediction Heads} \label{subsec:ffn}
The decoder network provides a set of embeddings where the embeddings serve as the latent representations for the action instances in the video. 
This output graph $\mathbf{y}_{L_d}$ contains $N_o$ nodes. We use these $N_o$ node embeddings to obtain predictions for $N_o$ action instances using prediction heads.  
The prediction heads consist of a feed forward network (FFN) with ReLU activation which provides the start time and end time of the action instance normalized with respect to the overall video duration. Additionally, we use a linear layer with a softmax function to predict the categorical label corresponding to the action instance. 

Therefore, when provided with the $i$-th node embedding $\mathbf{y}_{L_d}^{(i)}$, the prediction head provides prediction $\tilde a^{(i)} = (\tilde c^{(i)}, \tilde t^{(i)}_s, \tilde t^{(i)}_e)$ where $\tilde c^{(i)}$, $\tilde t^{(i)}_s$ and $\tilde t^{(i)}_e$ are the category label, start time and end time for the $i$-th action instance for $i \in \{1,2,\ldots, N_o\}$. Note that the ground truth set  would contain a variable number of action instances, whereas $N_o$ is larger than the maximum number of action instances per video in the dataset. This calls for a need to suppress irrelevant predictions. We do this by introducing an additional class label $\varnothing$ indicating no action (similar to ~\cite{carion2020end}). As such, this non-maximal suppression (typically performed using heuristics in existing methods~\cite{chao2018rethinking}) is learnable in our model.

\subsubsection{Loss functions}
\label{subsec:loss}
To train the overall network, we align the predictions with the ground truth action instances using a matcher module which optimizes a pair-wise cost function. This provides a unique matching between the predicted and ground truth action instances. Subsequently, our model computes losses corresponding to these matched pairs of predicted and ground truth action instances to train the overall network end-to-end.

\vspace{0.05in}
\noindent
\textbf{Matcher. } The matcher module finds an optimal matching between the predicted set of action instances (that contains fixed number of elements for every video) and the ground truth set of action instances (that contains a variable number of elements depending on the video). To obtain this matching, we design a matching cost function and employ the Hungarian algorithm to obtain an optimal matching between the two sets as described in prior work \cite{stewart2016end}.

Formally, let $\mathcal{A}$ be the ground truth set of action instances $\mathcal{A} = \{a^{(i)}\}^{|\mathcal{A}|}_{i=1}$, where $a^{(i)}= (c^{(i)}, t^{(i)}_s, t^{(i)}_e)$ and $\mathcal{\tilde A}$ be the predicted set of action instances $\mathcal{\tilde A} = \{\tilde a^{(i)}\}^{N_o}_{i=1}$ where $ \tilde a^{(i)}= (\tilde c^{(i)}, \tilde t^{(i)}_s,  \tilde t^{(i)}_e)$. 
In our model, we assume that $N_o$ is larger than the number of actions in any video in the dataset. Therefore, we assume that ground truth set $\mathcal{A}$ also is a set of size $N_o$ by padding the remaining $(N_o - |\mathcal{A}|)$ elements with $\varnothing$ element indicating no action. The optimal bipartite matching between the two sets reduces to choosing the permutation of $N_o$ elements $\hat \phi$ from the set of all possible permutations $\Phi_{N_o}$ that results in lowest value of the matching cost function $\mathcal{L}_{m}$. 
Thus, 
$ \hat \phi = \operatorname*{argmin}_{\phi \in \Phi_{N_o}} \mathcal{L}_{m}(a^{(i)},\tilde a^{(\phi(i))})$,
where $\mathcal{L}_{m}(a^{(i)},\tilde a^{(\phi(i))})$ is the matching cost function between ground truth $a^{(i)}$ and prediction with index $\phi(i)$. The matching cost function
incorporates the class probabilities of the action instances and the proximity between predicted and ground truth timestamps. Specifically, we define the cost function as:
\begin{equation}
\begin{split}
    \mathcal{L}_{m}(a^{(i)},\tilde a^{(\phi(i))}) & = - \mathbbm{1}_{\{c^{(i)} \neq \varnothing\}} \tilde p_{\phi(i)}(c^{(i)})  \\
    & + \mathbbm{1}_{\{c^{(i)} \neq \varnothing\}} \mathcal{L}_s(s^{(i)}, \tilde  s^{(\phi(i))}),
    \end{split}
\end{equation} 
where $s^{(i)} = [t^{(i)}_s, t^{(i)}_e]$ and $\tilde s^{(\phi(i))} = [\tilde t^{(\phi(i))}_s, \tilde t^{(\phi(i))}_e]$, and
$\tilde p_{\phi(i)}(c^{(i)})$ is the probability of class $c^{(i)}$ for prediction $\phi(i)$ and $\mathcal{L}_{s}$ represents segment loss that measures proximity in the timestamps of the instances. 
The segment loss is defined as a weighted combination of an $L_1$ loss (sensitive to the durations of the instances) and an IoU loss (invariant to the durations of the instances) between the predicted and ground-truth start and end timestamps. It is expressed as:
\begin{equation}
    \mathcal{L}_s = \lambda_{iou} \mathcal{L}_{iou}(s^{(i)}, \tilde s^{(\phi(i))})
     + \lambda_{L1} ||s^{(i)} - \tilde s^{(\phi(i))}||_{1} ,
\end{equation}
where $\lambda_{iou}, \lambda_{L1} \in \mathbbm{R}$ are hyperparameters. Subsequent to obtaining the optimal permutation $\hat \phi$, we compute the Hungarian loss $\mathcal{L}_H$ over all the matched pairs as follows:
\begin{equation}
\label{eq:loss}
\mathcal{L}_H = \sum_{i=1}^{N_o} \Big[-\log \tilde p_{\hat\phi} (c^{(i)})
 + \mathbbm{1}_{\{c^{(i)} \neq \varnothing\}} \mathcal{L}_s(s^{(i)},  \tilde s^{(\hat\phi(i))})\Big].
\end{equation}
This loss is used to train our AGT model end-to-end. We provide further implementation details of the model in the supplementary material.

In summary, our proposed Activity Graph Transformer performs temporal action localization using an encoder-decoder based architecture leveraging graph based attention modules. We jointly optimize all parameters of our model to minimize the regression loss for the start and end timestamps of the action instances and the cross entropy losses for the corresponding action labels.  


\section{Experiments}
We conducted several experiments to demonstrate the effectiveness of our proposed approach. In this section, we report the results of our evaluation.

\vspace{0.05in}
\noindent
\textbf{Datasets.} We use three benchmark datasets for evaluation. They vary in their extent of overlap in action instances, the number of action instances per video, and the number of action categories in the dataset. Thus, these datasets together serve as a challenging testbed for our model.

\textbf{THUMOS14}~\cite{THUMOS14} contains 200 videos in training set and 213 videos in testing set for the the task of action localization. This dataset has 20 action categories. The videos contain an average of 15 action instances per video with an average of 8\%  overlapping with other instances.

\textbf{Charades}~\cite{sigurdsson2016hollywood} is large scale dataset containing 9848 videos of daily indoor activities. This dataset has 157 action categories. Videos in the dataset contain an average of 6 action instances per video with an average of 79\% of overlapping instances in a video. This dataset is challenging because of the high degree of overlap in the action instances.

\textbf{EPIC-Kitchens100}~\cite{Damen2020RESCALING} contains 700 egocentric videos of daily kitchen activities. This dataset contains 289 noun and 97 verb classes. Videos in the dataset contain an average of 128 action instances per video with an average of 28\% overlapping instances in a video. 


\setlength{\tabcolsep}{4pt}
\renewcommand{\arraystretch}{0.95}
\begin{table}[t]
\centering
\caption{\textbf{Comparison with state-of-the-art (THUMOS14).} We report the mean average precision at different intersection over union thresholds (mAP@tIoU) for tIoU$\in \{0.1, 0.2, 0.3, 0.4, 0.5\}$. $\uparrow$ indicates higher is better.}
\scalebox{1.0}{
\begin{tabular}{l@{\hskip 3mm}c@{\hskip 4mm}c@{\hskip 4mm}c@{\hskip 4mm}c@{\hskip 4mm}c@{\hskip 4mm}c}
\toprule
\multirow{2}{*}{Method} & \multicolumn{5}{c}{mAP@tIoU $\uparrow$} \\
\cmidrule(lr){2-6}
& 0.1 & 0.2 & 0.3 & 0.4 & 0.5 \\
\midrule
Oneata \etal ~\cite{oneata2013action} & 36.6 & 33.6 & 27.0 & 20.8 & 14.4 \\
Wang \etal ~\cite{wang2014action} & 18.2 & 17.0 & 14.0 & 11.7 & \phantom{0}8.3 \\
Caba \etal ~\cite{caba2016fast} & - & - & - & - & 13.5 \\
Richard \etal ~\cite{richard2016temporal} & 39.7 & 35.7 & 30.0 & 23.2 & 15.2 \\
Shou \etal ~\cite{shou2016temporal} & 47.7 & 43.5 & 36.3 & 28.7 & 19.0 \\
Yeung \etal ~\cite{yeung2016end} & 48.9 & 44.0 & 36.0 & 26.4 & 17.1 \\
Yuan \etal ~\cite{yuan2016temporal} & 51.4 & 42.6 & 33.6 & 26.1 & 18.8 \\
Buch \etal ~\cite{buch2017sst} & - & - & 37.8 & - & 23.0 \\
Shou \etal ~\cite{shou2017cdc} & - & - & 40.1 & 29.4 & 23.3 \\
Yuan \etal ~\cite{yuan2017temporal} & 51.0 & 45.2 & 36.5 & 27.8 & 17.8 \\
Buch \etal ~\cite{buch2017end} & - & - & 45.7 & - & 29.2 \\
Gao \etal ~\cite{gao2017turn} & 60.1 & 56.7 & 50.1 & 41.3 & 31.0 \\
Dai \etal ~\cite{dai2017temporal} & - & - & - & 33.3 & 25.6 \\
Xu \etal ~\cite{xu2017r} & 54.5 & 51.5 & 44.8 & 35.6 & 28.9 \\
Zhao \etal ~\cite{zhao2017temporal} & 66.0 & 59.4 & 51.9 & 41.0 & 29.8 \\
Lin \etal ~\cite{lin2018bsn} & - & - & 53.5 & 45.0 & 36.9 \\
Chao \etal ~\cite{chao2018rethinking} & 59.8 & 57.1 & 53.2 & 48.5 & 42.8 \\
Zeng \etal ~\cite{zeng2019graph} & 69.5 & 67.8 & 63.6 & 57.8 & 49.1 \\
Xu \etal ~\cite{xu2020g} & 66.1 & 64.2 & 54.5 & 47.6 & 40.2\\
\midrule
\textbf{AGT (Ours)} & \textbf{72.1} & \textbf{69.8} & \textbf{65.0} & \textbf{58.1} & \textbf{50.2} \\

\bottomrule
\end{tabular}}
\label{tab:thumos_sota}
\end{table}

\vspace{0.05in}
\noindent
\textbf{Comparison with state-of-the-art. } We compare the performance of our proposed AGT with the state-of-the-art methods. We use mean average precision as the metric to evaluate the model. To ensure fair comparison, we use the same evaluation protocol as used by state-of-the-art methods for each of the datasets. Table~\ref{tab:thumos_sota} shows that the our AGT achieves upto 3.5\% improvement over state-of-the-art for THUMOS14 dataset and consistently shows performance improvement  across all IoU thresholds. Table~\ref{tab:charades_sota} shows the comparisons with state-of-the-art methods on Charades dataset. Our model achieves 13\% improvement in the Charades dataset. We also perform comparison on recenty released EPIC-Kitchens100 dataset for classification of verb, noun, and action (\ie both verb and noun) classes. Table~\ref{tab:epic_sota} indicates that our model performs consistently for all three tasks for EPIC-Kitchens100 datasets across all IoU thresholds. Overall, these results clearly show that our proposed method AGT outperforms the state-of-the-art methods by a considerable margin. 
\setlength{\tabcolsep}{4pt}
\renewcommand{\arraystretch}{0.95}
\begin{table}[t]
\centering
\caption{\textbf{Comparison with state-of-the-art (Charades).} We report mean average precision (mAP) computed using \texttt{Charades\_v1\_localize} setting in ~\cite{sigurdsson2016hollywood}. $\uparrow$: higher is better.}
\scalebox{0.9}{
\begin{tabular}{l c}
\toprule
Method & mAP $\uparrow$ \\
\midrule
Predictive-corrective (Dave \etal ~\cite{dave2017predictive}) & \phantom{0}8.9 \\
Two-stream (Siggurdson \etal ~\cite{sigurdsson2016hollywood}) & \phantom{0}8.9 \\
Two-stream + LSTM (Siggurdson \etal ~\cite{sigurdsson2016hollywood}) & \phantom{0}9.6 \\
R-C3D (Xu \etal ~\cite{xu2017r}) & 12.7 \\
SSN (Zhao \etal ~\cite{zhao2017temporal}) & 16.4 \\
I3D baseline ~\cite{piergiovanni2019temporal} & 17.2 \\
Super-events (Piergiovanni \etal ~\cite{piergiovanni2018learning}) & 19.4 \\
TGM (Piergiovanni \etal ~\cite{piergiovanni2018learning}) & 22.3 \\
Mavroudi \etal ~\cite{mavroudi2020representation} & 23.7 \\
3D ResNet-50 + super-events (Piergiovanni \etal ~\cite{piergiovanni2020avid}) & 25.2\\

\midrule
\textbf{AGT (Ours)} & \textbf{28.6}\\
\bottomrule
\end{tabular}}
\label{tab:charades_sota}
\end{table}

\setlength{\tabcolsep}{4pt}
\renewcommand{\arraystretch}{0.95}
\begin{table}[t]
\centering
\caption{\textbf{Comparison with state-of-the-art (EPIC-Kitchens100).} We report mean average precision at different intersection over union thresholds (mAP@tIoU) for tIoU$\in \{0.1, 0.2, 0.3, 0.4, 0.5\}$. We use the validation split in the original dataset for testing. $\uparrow$ indicates higher is better.}

\scalebox{0.91}{
\begin{tabular}{l@{\hskip 0.2cm}l@{\hskip 2mm}c@{\hskip 4mm}c@{\hskip 4mm}c@{\hskip 4mm}c@{\hskip 4mm}c@{\hskip 4mm}c}
\toprule
& \multirow{2}{*}{Method} & \multirow{2}{*}{Task} & \multicolumn{5}{c}{mAP@tIoU $\uparrow$} \\
\cmidrule(lr){4-8}
& & & 0.1 & 0.2 & 0.3 & 0.4 & 0.5 \\
\midrule
&  & Verb & 10.51 & \phantom{0}9.24 & \phantom{0}7.67 & \phantom{0}6.40 & \phantom{0}5.12 \\
& Damen  & Noun & 10.71 & \phantom{0}8.73 & \phantom{0}6.75 & \phantom{0}5.05 & \phantom{0}3.35 \\
& \etal ~\cite{Damen2020RESCALING}&Action & \phantom{0}6.78 & \phantom{0}6.03 & \phantom{0}4.94 & \phantom{0}4.04 & \phantom{0}3.35 \\

\midrule
&  & Verb & 12.01 & 10.25 & \phantom{0}8.15 & \phantom{0}7.12 & \phantom{0}6.14 \\
& \textbf{AGT} & Noun & 11.63 & \phantom{0}9.33 & \phantom{0}7.05 & \phantom{0}6.57 & \phantom{0}3.89 \\
&  \textbf{(Ours)} &Action & \phantom{0}7.78 & \phantom{0}6.92 & \phantom{0}5.53 & \phantom{0}4.22 & \phantom{0}3.86\\
\bottomrule
\end{tabular}}
\label{tab:epic_sota}
\end{table}


\vspace{0.05in}
\noindent
\textbf{Impact of graph based reasoning. }To demonstrate the importance of reasoning over videos as graphs, we conducted ablation studies by removing the graph based reasoning components from either the encoder or the decoder or both (\ie overall transformer network) in our model. Specifically, this is implemented by removing the graph message passing layers from the attention modules (\ie, graph self-attention module and graph-to-graph attention module) in the encoder and/or decoder blocks in the network. Intuitively, when the graph message passing module is removed from the whole transformer network, the transformer encoder treats the input as a sequence and the transformer decoder treats the action queries as independent. Table~\ref{tab:ablation_structure} shows the performance of these ablated versions of our model. The results clearly show that eliminating the graph-based reasoning module hurts the localization performance. The results also suggest that graph-based modeling is more useful in the encoder than in the decoder. We believe this is because the graph reasoning performed by the encoder is more useful in capturing the non-sequential dependencies as it operates directly on the video features.
For better readability, here, we provide the mAP values averaged over the various intersection-over-union thresholds (tIoU) for THUMOS14 and EPIC-Kitchens100. For mAP values at specific thresholds, refer to the supplementary.
\setlength{\tabcolsep}{4pt}
\renewcommand{\arraystretch}{1}
\begin{table}[t]
\centering
\caption{\textbf{ Ablation Study (Impact of graph based reasoning).} We report performance of ablated versions of our AGT model. We report mAP for evaluation performance (higher is better). We remove graph reasoning in the encoder ($\mathbf{E}$) and/or decoder ($\mathbf{D}$) of the transformer. \cmark\ and \xmark\ indicates whether a component (encoder or decoder) contains graph message passing module or not respectively. 
EPIC(A), EPIC (V), EPIC (N) indicates task `Action', `Verb', `Noun' classification on EPIC-Kitchens100. }
\scalebox{0.82}{
\begin{tabular}{l@{\hskip 3mm}c@{\hskip 4mm}c@{\hskip 4mm}c@{\hskip 4mm}c}
\toprule
Dataset & $\mathbf{E}$: \xmark/ $\mathbf{D}$: \xmark & $\mathbf{E}$: \xmark/ $\mathbf{D}$: \cmark & $\mathbf{E}$: \cmark/ $\mathbf{D}$: \xmark & $\mathbf{E}$: \cmark/ $\mathbf{D}$: \cmark \\
\midrule
THUMOS14 & 55.6& 56.3& 58.3& 63.0\\
Charades & 18.2& 19.2&22.5&28.6\\
EPIC (A) & \phantom{0}3.0&\phantom{0}3.3 &\phantom{0}5.1&\phantom{0}5.9\\
EPIC (V) &\phantom{0}5.7&\phantom{0}6.1&\phantom{0}7.4 & \phantom{0}8.7\\
EPIC (N) &\phantom{0}4.9&\phantom{0}5.3&\phantom{0}6.3 & \phantom{0}7.7\\

\bottomrule
\end{tabular}}
\vspace{-1mm}
\label{tab:ablation_structure}
\end{table}

\setlength{\tabcolsep}{4pt}
\renewcommand{\arraystretch}{0.95}
\begin{table}[t]
\centering
\caption{\textbf{ Ablation Study (Impact of temporal resolution).} Performance of our AGT for different temporal resolutions of input video. Here, SR indicates sampling rate of frames for feature extraction, \ie, SR=1/$k$ means frames sampled at 1/$k$ -th factor of the original frame rate. EPIC(A), EPIC (V), EPIC (N) indicate tasks `Action', `Verb', `Noun' on dataset EPIC-Kitchens100. We report mAP for evaluation (higher is better).}
\scalebox{0.85}{
\begin{tabular}{l@{\hskip 10mm}c@{\hskip 10mm}c@{\hskip 10mm}c}
\toprule
Dataset & SR = 1/8 & SR=1/4 \\
\midrule
THUMOS14 & 60.2 & 63.0\\
Charades & 27.3 & 28.6\\
EPIC (A) & 4.1 & 5.9\\
EPIC (V) & 7.2& 8.7\\
EPIC (N) & 6.4& 7.7\\
\bottomrule
\end{tabular}}
\vspace{-4mm}
\label{tab:ablation_tempres}
\end{table}


\vspace{0.05in}
\noindent
\textbf{Impact of temporal resolution.} To evaluate the impact of temporal resolution, we experimented with different frame rates for the input video. Table~\ref{tab:ablation_tempres} shows the results suggesting higher resolution leads to better performance as the higher temporal resolution provides more information in the input. However, our results also show that lower resolution does not lead to any major drop in performance. For better readability, here, we provide the mAP values averaged over the various intersection-over-union thresholds (tIoU) for THUMOS14 and EPIC-Kitchens100. mAP values at specific thresholds are available in the supplementary.

\begin{figure}[!t]
    \centering
    \begin{tabular}{c}
        \includegraphics[width=0.46\textwidth]{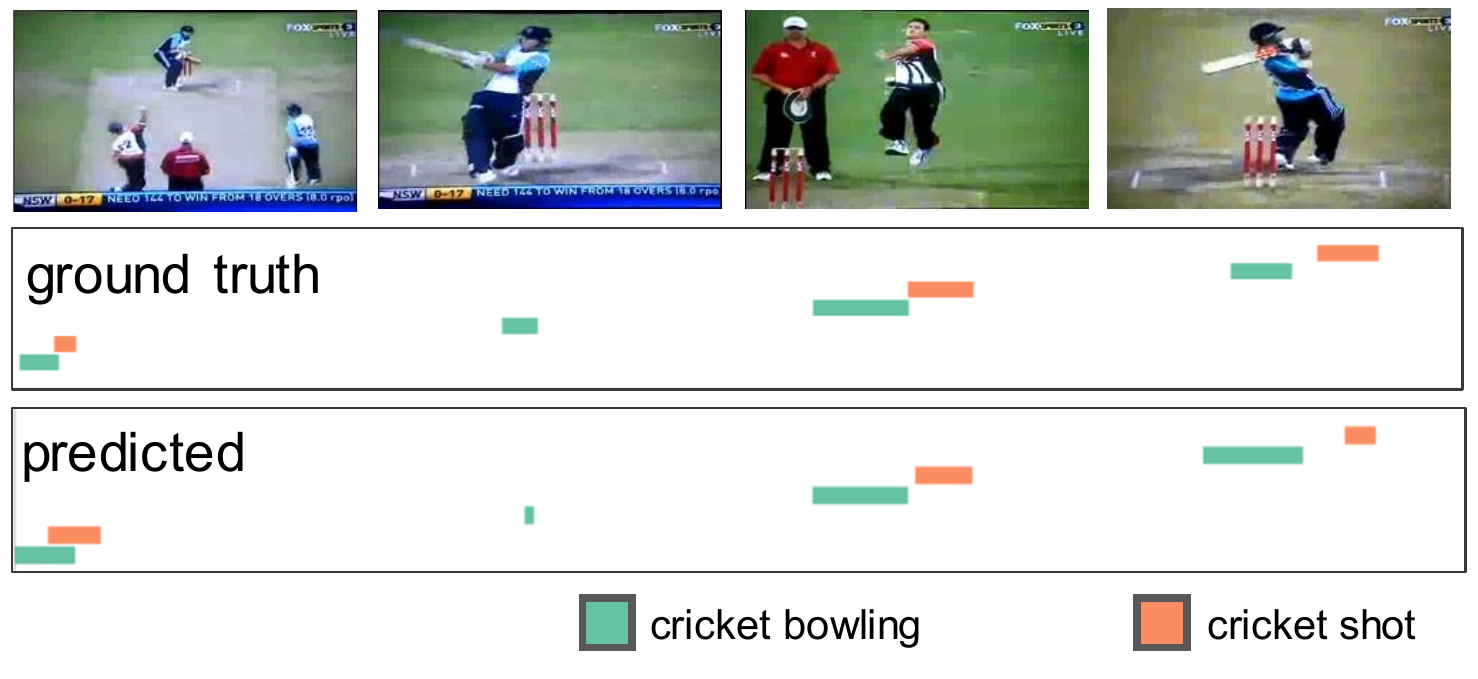}\\
        \includegraphics[width=0.46\textwidth]{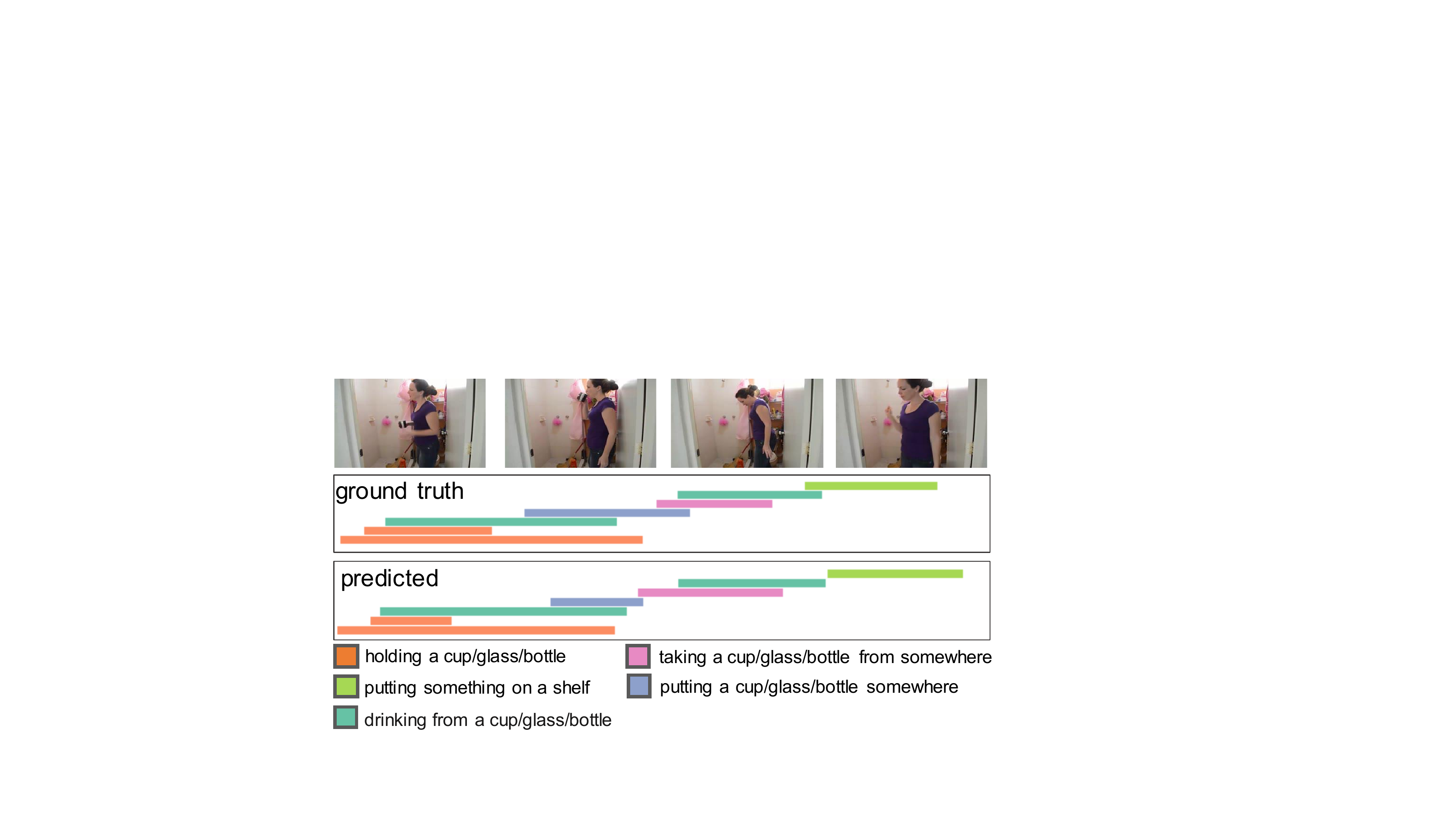}
    \end{tabular}

    \vspace{-1mm}
    \caption{\textbf{Qualitative Results. }Visualization of ground truth and predicted action instances.}
    \vspace{-5mm}
    \label{fig:qual}
\end{figure}

\vspace{0.05in}
\noindent 
\textbf{Qualitative Results.}
We visualize the predictions of the model on two different samples in Figure~\ref{fig:qual}. The visualizations indicate that our model is able to predict the correct number of action instances as well as correct action categories with minimal errors in start and end timestamps. We believe this is because video content around the start and end timestamps in some instances do not contain enough information pertaining to the action. We provide additional visualizations of  predictions in the supplementary. 

Additionally, refer to the supplementary for experiments on performance of our model with varied number of layers and heads in the transformer and ablations of loss functions.





\section{Conclusion}
In this paper, we proposed a novel end-to-end learnable encoder-decoder transformer model for the task of temporal action localization in untrimmed human activity videos. Our approach aims to model the non-linear temporal structure in such videos by reasoning over the videos as graphs using graph self-attention mechanisms. The experimental evaluation showed that our model achieves state-of-the-art performance on the task of temporal action localization on challenging human activity datasets. Overall, this work highlights the importance of reasoning over videos as non-sequential entities and shows that graph-based transformers are an effective means to model complex activity videos.
\clearpage

{\small
\bibliographystyle{ieee_fullname}
\bibliography{paper_review}
}

\clearpage
\cvprappendix

\renewcommand\thesection{\Alph{section}}
\setcounter{table}{0}
\setcounter{figure}{0}
\setcounter{section}{0}
\renewcommand{\thetable}{T\arabic{table}}
\renewcommand{\thesection}{\Alph{section}}
\renewcommand{\thefigure}{F\arabic{figure}}

\section{Appendix}
We report additional quantitative results and qualitative analysis and provide implementation details of our model. Specifically, this document contains the following.
\begin{itemize}
    \item Code provided in the folder \texttt{agt\_code.zip}
    
    \item Additional quantitative evaluation
     \vspace{-1mm}
     \begin{itemize}
     \setlength\itemsep{0mm}
     \item Section~\ref{subsec:eval_supp}: Supplemental tables for Table 4 and Table 5 from the main paper to report mAP values at specific IoU thresholds
     \item Section~\ref{subsec:ablation_loss}: Ablation study of loss function (Eq. 9 in the main paper).
     \item  Section~\ref{subsec:ablation_layers}: Impact of different number of layers in transformer encoder and decoder.
     \item Section~\ref{subsec:ablation_heads}: Impact of different number of heads in attention modules of the transformer.
     \item Section~\ref{subsec:ablation_queries}: Impact of different number of nodes in the action query graph.
     \end{itemize}
     \item Additional qualitative analysis
    \vspace{-1mm}
     \begin{itemize}
        \setlength\itemsep{0mm}
         \item Section~\ref{subsec:vis_preds}: Visualization of predictions
         \item Section~\ref{subsec:vis_graphs}: Visualization of graphs learned by the model
         \item
        Section~\ref{subsec:vis_analysis}: Analysis of AGT predictions based on the duration of action instances 
         
     \end{itemize}
     \item Technical details
     \vspace{-1mm}
     \begin{itemize}
        \setlength\itemsep{0mm}
         \item Section~\ref{subsec:imp_arch}: Details of the architecture of AGT
         \item Section~\ref{subsec:imp_tech}: Details of initialization, data augmentation, and hyperparameters.
     \end{itemize}

\end{itemize}

\subsection{Additional Quantitative Evaluation}
In this section, we report the quantitative evaluation of our proposed AGT model to supplement the quantitative evaluation in the main paper.

\subsubsection{Supplemental Tables}
\label{subsec:eval_supp}
In the main paper, we only reported the mAP averaged over different IoU thresholds for THUMOS14 and EPIC-Kitchens100 dataset (Table 4 and Table 5 in main paper). For completeness, we report mAP at specific IoU thresholds in Table~\ref{tab:supp_tab4} and Table~\ref{tab:supp_tab5}.

\setlength{\tabcolsep}{4pt}
\renewcommand{\arraystretch}{1}
\begin{table}[t]
\centering
\caption{\small{\textbf{Supplemental Tables: Impact of graph based reasoning.} We report performance of ablated versions of our AGT model. We remove graph reasoning in the encoder ($\mathbf{E}$) and/or decoder ($\mathbf{D}$) of the transformer. \cmark\ and \xmark\ indicates whether a component (encoder or decoder) contains graph message passing module or not respectively. 
EPIC (A), EPIC (V), EPIC (N) indicates task `Action', `Verb', `Noun' classification on EPIC-Kitchens100. We report the mean average precision at different intersection over union thresholds (mAP@tIoU) for tIoU$\in \{0.1, 0.2, 0.3, 0.4, 0.5\}$. $\uparrow$ indicates higher is better.}}

\scalebox{0.9}{
\begin{tabular}{c@{\hskip 4mm}c@{\hskip 4mm}c@{\hskip 4mm}c@{\hskip 4mm}c@{\hskip 4mm}c@{\hskip 4mm}c}
\toprule
\multirow{2}{*}{Dataset} & \multirow{2}{*}{Model} & \multicolumn{5}{c}{mAP@tIoU $\uparrow$} \\
\cmidrule(lr){3-7}
& & 0.1 & 0.2 & 0.3 & 0.4 & 0.5 \\
\midrule
 & $\mathbf{E}$: \xmark/ $\mathbf{D}$: \xmark & 64.6 & 60.8 & 59.1 & 51.2& 40.3\\
THUMOS14 &$\mathbf{E}$: \xmark/ $\mathbf{D}$: \cmark & 65.1 & 62.4 & 60.3 & 52.4 & 41.3\\
 & $\mathbf{E}$: \cmark/ $\mathbf{D}$: \xmark & 67.1 & 64.4 & 62.5 & 53.6 & 44.9\\
 & $\mathbf{E}$: \cmark/ $\mathbf{D}$: \cmark & 72.1 & 69.8 & 65.0 & 58.1 & 50.2\\
\midrule
 & $\mathbf{E}$: \xmark/ $\mathbf{D}$: \xmark & 9.4& 6.9& 5.2& 4.5& 2.5\\
EPIC (V) &$\mathbf{E}$: \xmark/ $\mathbf{D}$: \cmark & 9.9 & 7.5 & 5.5 & 4.9 & 2.7\\
 & $\mathbf{E}$: \cmark/ $\mathbf{D}$: \xmark & 11.4 & 9.0 & 6.9 & 6.3 & 3.4\\
 & $\mathbf{E}$: \cmark/ $\mathbf{D}$: \cmark & 12.0 & 10.3 & 8.2 & 7.1 & 6.1\\
\midrule
 & $\mathbf{E}$: \xmark/ $\mathbf{D}$: \xmark & 8.9 & 5.4 & 4.9 & 3.6 & 1.7\\
EPIC (N) &$\mathbf{E}$: \xmark/ $\mathbf{D}$: \cmark & 9.2 & 6.0 & 5.1 & 4.2 & 2.0\\
 & $\mathbf{E}$: \cmark/ $\mathbf{D}$: \xmark & 10.1 & 8.0 & 6.8 & 5.2 & 2.3\\
 & $\mathbf{E}$: \cmark/ $\mathbf{D}$: \cmark & 11.6 & 9.3 & 7.1 & 6.6 & 3.9\\
\midrule
 & $\mathbf{E}$: \xmark/ $\mathbf{D}$: \xmark & 4.8 & 4.1 & 2.9 & 2.1 & 1.5\\
EPIC (A) &$\mathbf{E}$: \xmark/ $\mathbf{D}$: \cmark & 5.1 & 4.3 & 3.2 & 2.3 & 1.8\\
 & $\mathbf{E}$: \cmark/ $\mathbf{D}$: \xmark & 7.3 & 6.1 & 5.0 & 3.9 & 3.7\\
 & $\mathbf{E}$: \cmark/ $\mathbf{D}$: \cmark & 7.8 & 6.9 & 5.5 & 4.2 & 3.9\\
\bottomrule
\end{tabular}}
\vspace{-0.5cm}
\label{tab:supp_tab4}
\end{table}

\setlength{\tabcolsep}{4pt}
\renewcommand{\arraystretch}{1}
\begin{table}[t]
\centering
\caption{\small{\textbf{ Supplemental Tables: Impact of temporal resolution.} Performance of our AGT for different temporal resolutions of input video. Here, SR indicates sampling rate of frames for feature extraction, \ie, SR=1/$k$ means frames sampled at 1/$k$ -th factor of the original frame rate. EPIC (A), EPIC (V), EPIC (N) indicate tasks `Action', `Verb', `Noun' classification on dataset EPIC-Kitchens100. We report the mean average precision at different intersection over union thresholds (mAP@tIoU) for tIoU$\in \{0.1, 0.2, 0.3, 0.4, 0.5\}$. $\uparrow$ indicates higher is better.}}

\scalebox{0.9}{
\begin{tabular}{c@{\hskip 4mm}c@{\hskip 4mm}c@{\hskip 4mm}c@{\hskip 4mm}c@{\hskip 4mm}c@{\hskip 4mm}c}
\toprule
\multirow{2}{*}{Dataset} & \multirow{2}{*}{Model} & \multicolumn{5}{c}{mAP@tIoU $\uparrow$} \\
\cmidrule(lr){3-7}
& & 0.1 & 0.2 & 0.3 & 0.4 & 0.5 \\
\midrule
THUMOS14 & SR=1/8 & 70.4 & 65.8 & 62.3 & 54.1 & 48.6 \\
 & SR=1/4 & 72.1 & 69.8 & 65.0 & 58.1 & 50.2\\
\midrule
EPIC (V) & SR=1/8 & 10.9 & 8.5 & 6.3 & 5.5 & 4.9\\
 & SR=1/4 & 12.0 & 10.3 & 8.2 & 7.1 & 6.1\\
\midrule
EPIC (N) &SR=1/8 & 10.2 & 8.1 & 5.9 & 5.2 & 2.8 \\
 & SR=1/4 & 11.6 & 9.3 & 7.1 & 6.6 & 3.9\\
\midrule
EPIC (A) & SR=1/8 & 7.0 & 5.0 & 4.1 & 2.5 & 1.9 \\
 & SR=1/4 & 7.8 & 6.9 & 5.5 & 4.2 & 3.9 \\
\bottomrule
\end{tabular}}
\label{tab:supp_tab5}
\end{table}

\subsubsection{Ablation Study: Loss function}
\label{subsec:ablation_loss}
Note that for any version of the loss function, the model requires cross entropy loss to be able to classify the action label pertaining to an instance. The model also requires some form of regression loss to produce predictions pertaining to the start and end timestamps of an action instance. Recall, our overall loss (see Eq. (9) in the main paper) is a combination of cross-entropy loss and regression loss, \ie, segment loss $\mathcal{L}_s$. The segment loss contains two components: $L_1$ loss and IoU loss $\mathcal{L}_{iou}$. Table~\ref{tab:ablation_loss} shows the results of the performance of our model when trained with ablated versions of the segment loss. The results indicate that the models trained with only $L_1$ loss perform better than the ones trained with only IoU loss $\mathcal{L}_{iou}$. Additionally, models trained with both losses are better than the ones trained with only one of the losses. Nonetheless, all versions of our AGT model perform better than state-of-the-art methods. We only provide the mAP values averaged over the various intersection-over-union thresholds (tIoU) for THUMOS14 and EPIC-Kitchens100.
\setlength{\tabcolsep}{4pt}
\renewcommand{\arraystretch}{1}
\begin{table}[t]
\centering
\caption{\textbf{Ablation Study: Loss function.} We report performance of our AGT model when trained with ablated versions of the loss function. We report mAP for evaluation performance (higher is better). We train the model with a combination of cross-entropy loss and segment loss containing $L_1$ loss and/or IoU loss $\mathcal{L}_{iou}$. \cmark\ and \xmark\ indicate whether the specific component of the segment loss is used or not respectively. 
EPIC(A), EPIC (V), EPIC (N) indicate tasks `Action', `Verb', `Noun' classification on EPIC-Kitchens100. }
\scalebox{1.0}{
\begin{tabular}{l@{\hskip 6mm}c@{\hskip 6mm}c@{\hskip 6mm}c}
\toprule
Dataset & $L_1$: \cmark & $L_1$: \xmark & $L_1$: \cmark \\
 & $\mathcal{L}_{iou}$: \xmark &  $\mathcal{L}_{iou}$: \cmark &  $\mathcal{L}_{iou}$: \cmark \\
\midrule
THUMOS14 & 61.3 & 59.6 & 63.0\\
Charades & 26.0 & 25.3 & 28.6\\
EPIC (A) & \phantom{0}4.8& \phantom{0}3.7&\phantom{0}5.9\\
EPIC (V) & \phantom{0}7.1& \phantom{0}6.4 & \phantom{0}8.7\\
EPIC (N) & \phantom{0}6.3 & \phantom{0}5.0 & \phantom{0}7.7\\

\bottomrule
\end{tabular}}
\vspace{-1mm}
\label{tab:ablation_loss}
\end{table}

\subsubsection{Impact of number of layers}
\label{subsec:ablation_layers}
Table~\ref{tab:ablation_layers} shows the results of the performance of our model with different number of layers in encoder and decoder component of the transformer. While increase in number of layers increases the training time, we did not observe much difference in the performance of the model with increased depth of the transformer components. We only provide the mAP values averaged over the various intersection-over-union thresholds (tIoU) for THUMOS14 and EPIC-Kitchens100.
\setlength{\tabcolsep}{4pt}
\renewcommand{\arraystretch}{1}
\begin{table}[t]
\centering
\caption{\textbf{Impact of number of layers.} We report performance of our AGT model with different number of layers in encoder and decoder. We report mAP for evaluation performance (higher is better). EPIC (A), EPIC (V), EPIC (N) indicates task `Action', `Verb', `Noun' classification on EPIC-Kitchens100. \#E indicates number of layers in encoder and \#D indicates number of layers in decoder.}
\scalebox{1.0}{
\begin{tabular}{l@{\hskip 12mm}c@{\hskip 12mm}c@{\hskip 12mm}c}
\toprule
Dataset & \#E & \#D & mAP \\
\midrule
 & 4 & 2 & 62.5\\
THUMOS14 & 4 & 4 & 63.0\\
 & 2 & 4 & 62.7\\
\midrule
 & 4 & 2 & 28.0\\
Charades & 4 & 4 & 28.6\\
 & 2 & 4 & 28.2\\
\midrule
 & 4& 2& 5.5\\
EPIC (A) & 4 & 4 & 5.9\\
 & 2& 4& 5.6\\
\midrule
 & 4& 2& 8.5\\
EPIC (V) & 4& 4& 8.7\\
 & 2& 4& 8.6\\
\midrule
 & 4& 2& 7.2\\
EPIC (N) & 4& 4& 7.7\\
 & 2& 4& 7.3\\
\bottomrule
\end{tabular}}
\vspace{-1mm}
\label{tab:ablation_layers}
\end{table}

\subsubsection{Impact of number of heads}
\label{subsec:ablation_heads}
Table~\ref{tab:ablation_heads} shows the results of the performance of our AGT model with different number of heads in the attention modules of the transformer. The results suggest a slight improvement with more number of heads in the transformer network. We only provide the mAP values averaged over the various intersection-over-union thresholds (tIoU) for THUMOS14 and EPIC-Kitchens100.
\setlength{\tabcolsep}{4pt}
\renewcommand{\arraystretch}{1}
\begin{table}[t]
\centering
\caption{\textbf{Impact of number of heads.} We report performance of our AGT model with different number of heads in the attention modules of the transformer network. We report mAP for evaluation performance (higher is better). EPIC (A), EPIC (V), EPIC (N) indicates task `Action', `Verb', `Noun' classification on EPIC-Kitchens100. \#heads indicates number of heads in attention modules of the transformer.}
\scalebox{1.0}{
\begin{tabular}{l@{\hskip 12mm}c@{\hskip 12mm}c}
\toprule
Dataset & \#heads & mAP \\
\midrule
THUMOS14 & 8& 63.0\\
 & 4 & 61.4\\
\midrule
Charades &  8& 28.6\\
 & 4 & 26.4\\
\midrule
EPIC (A) & 8& 5.9\\
 & 4& 5.2\\
\midrule
EPIC (V) & 8& 8.7\\
 & 4& 8.4\\
\midrule
EPIC (N) & 8& 7.7\\
 & 4& 7.1\\
\bottomrule
\end{tabular}}
\vspace{-1mm}
\label{tab:ablation_heads}
\end{table}

\subsubsection{Impact of action query graph size}
\label{subsec:ablation_queries}
Table~\ref{tab:ablation_queries} shows the results of the performance of our AGT model with different number of node encodings in the action query graph. Intuitively, a very large size of action query graph implies the model will require more time to learn the non-maximal suppression of the irrelevant predictions. On the other hand, a very small size of action query graph might limit the ability of model to learn complex structure in the action instances. Note that, any value used for our experiments is higher than the maximum number of action instances per video in the dataset. The results suggest minor improvement with more number of nodes in the action query graph, however, the models with more number of nodes require longer training times. Our experiments also suggest that when the size of the action query graph is reduced, the localization performance of our model degrades. We only provide the mAP values averaged over the various intersection-over-union thresholds (tIoU) for THUMOS14 and EPIC-Kitchens100.
\setlength{\tabcolsep}{4pt}
\renewcommand{\arraystretch}{1}
\begin{table}[!t]
\centering
\caption{\textbf{Impact of action query graph size.} We report performance of our AGT model with different number of nodes in the action query graph. We report mAP for evaluation performance (higher is better). EPIC (A), EPIC (V), EPIC (N) indicates task `Action', `Verb', `Noun' classification on EPIC-Kitchens100. \#queries indicates number of nodes in the action query graph.}
\scalebox{1.0}{
\begin{tabular}{l@{\hskip 12mm}c@{\hskip 12mm}c}
\toprule
Dataset & \#queries & mAP \\
\midrule
  & 150 & 59.1\\
THUMOS14 & 300& 63.0\\
  & 900 & 63.2\\
\midrule
 & 30 & 24.0\\
Charades & 50 & 28.6\\
 & 100 & 28.6\\
\midrule
 & 900 & 4.3\\
EPIC (A) & 1500& 5.9\\
 & 2000 & 7.0\\
\midrule
 & 900 & 7.0\\
EPIC (V) & 1500& 8.7\\
 & 2000 & 8.8\\
\midrule
 & 900 & 6.1\\
EPIC (N) & 1500& 7.7\\
 & 2000 & 7.9\\
\bottomrule
\end{tabular}}
\vspace{-1mm}
\label{tab:ablation_queries}
\end{table}

\subsection{Additional Qualitative Analysis}
In this section, we visualize the results of our proposed model AGT to supplement the qualitative analysis in the main paper.

\subsubsection{Visualization: Predictions}
\label{subsec:vis_preds}
\begin{figure*}
\centering
    \begin{tabular}{c}
    \includegraphics[width=0.9\textwidth]{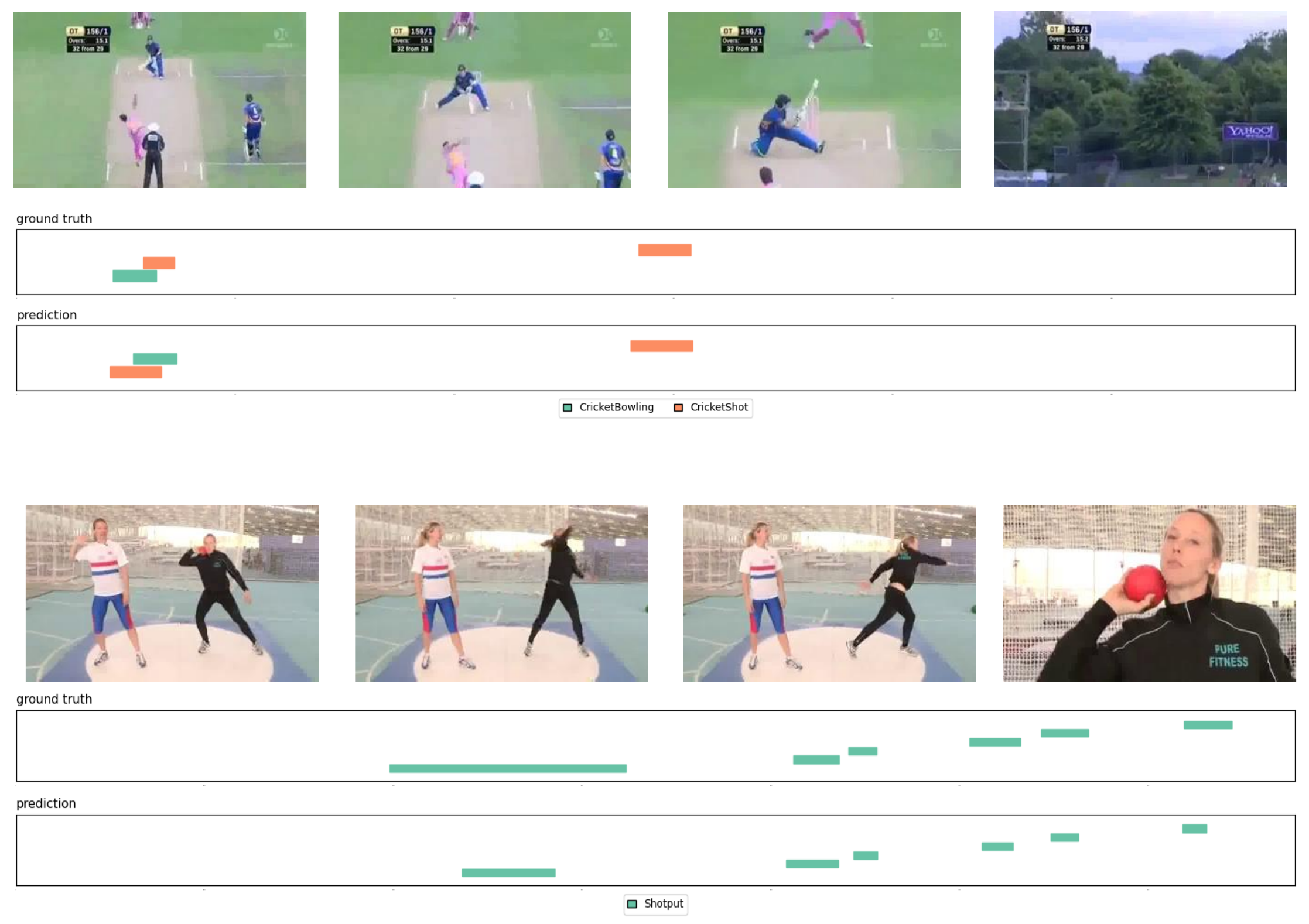}
    \\
    \includegraphics[width=0.9\textwidth]{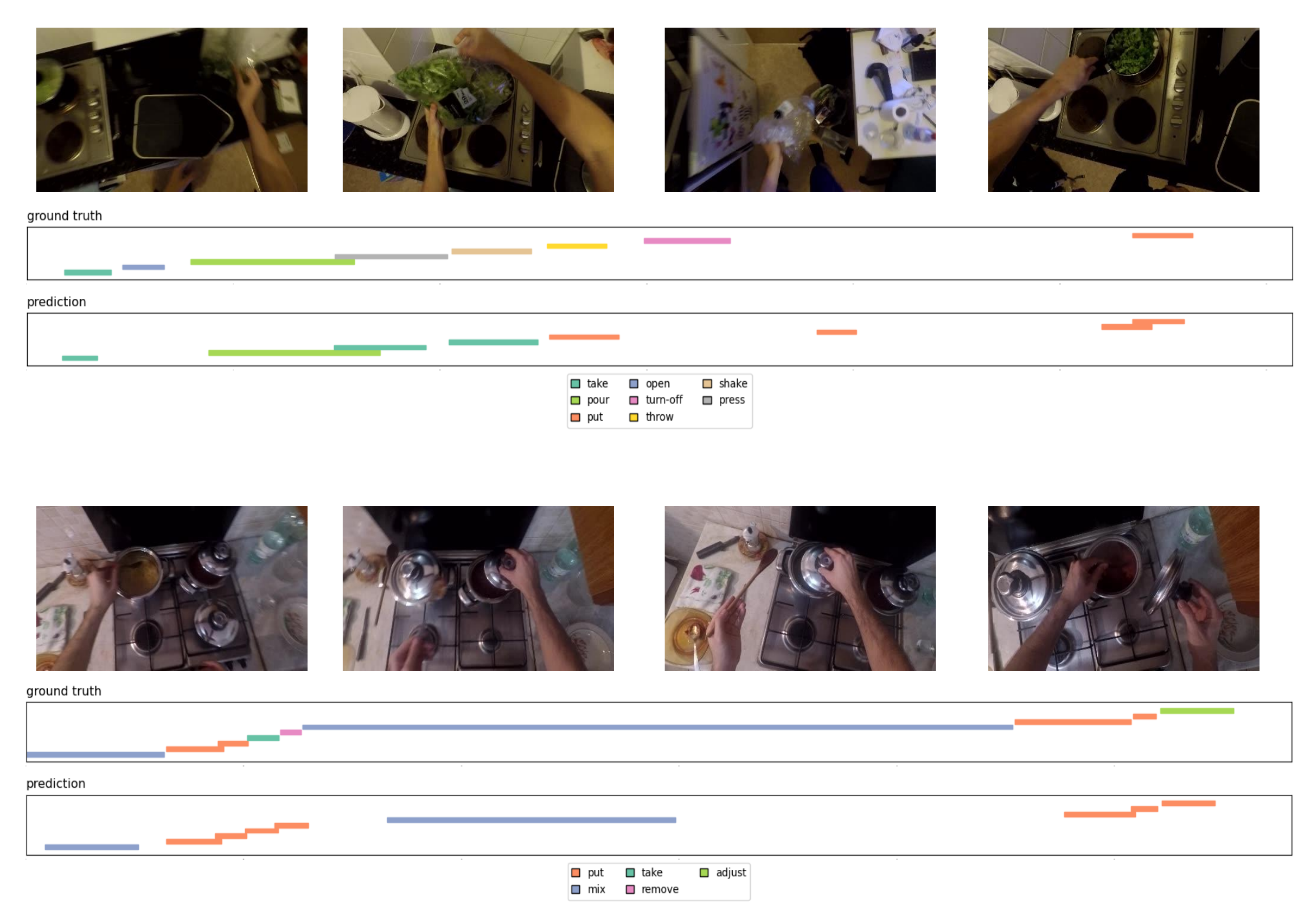}
    \end{tabular}
    \caption{\textbf{Visualization: Predictions. } Visualization of predictions and groundtruth action instances}
    \label{fig:vis_preds}
\end{figure*}
We provide additional visualizations of the predictions of our AGT on several diverse samples in Figure~\ref{fig:vis_preds}. The visualizations indicate that our model is able to predict the correct number of action instances as well as most of the correct action categories with minimal errors in start and end timestamps for videos containing overlapping instances with varying temporal extents. 

\subsubsection{Visualization: Learned Graphs}
\label{subsec:vis_graphs}
\begin{figure*}
    \begin{tabular}{c c}
        \includegraphics[width=0.45\textwidth]{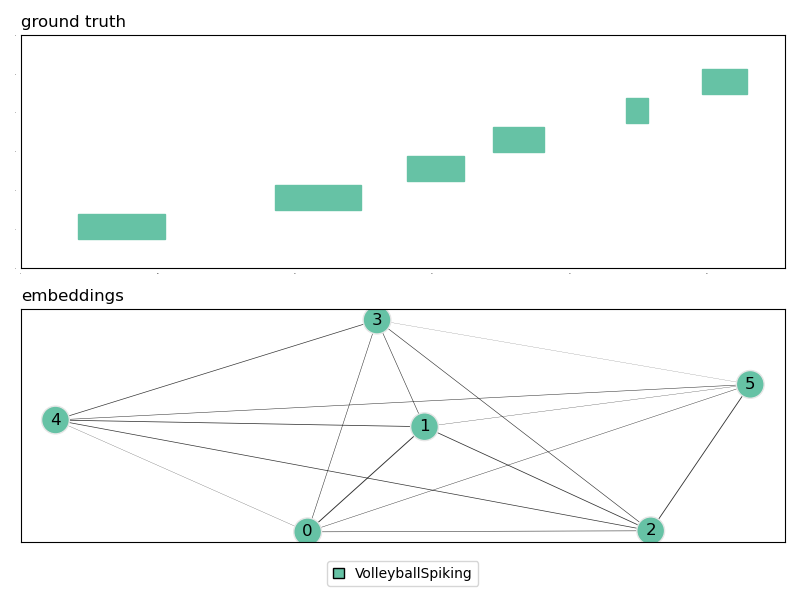}
         & 
         \includegraphics[width=0.45\textwidth]{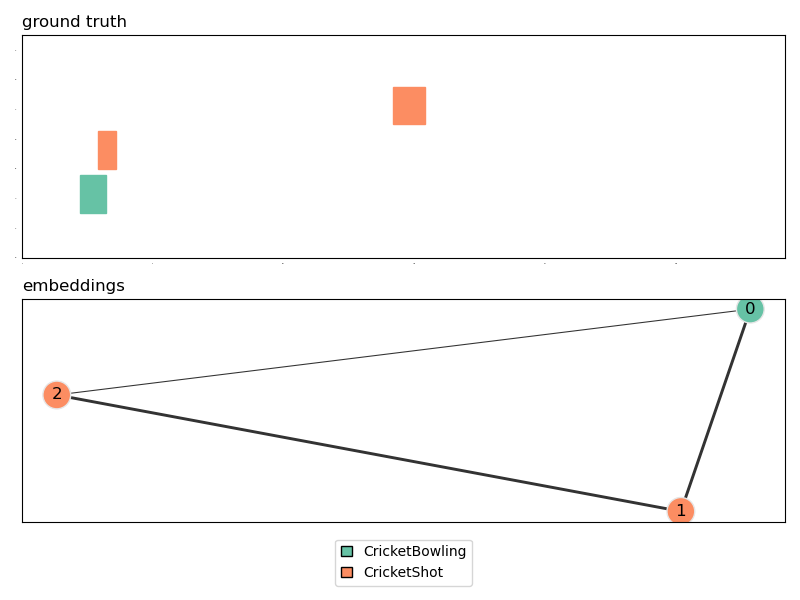}
           
    \end{tabular}
    \caption{\textbf{Visualization: Learned Graphs. } Visualizations of embeddings corresponding to the last layer of the decoder and ground truth instances. The thickness of edges show the strength of interaction between the nodes. For ease of visibility, the nodes have been numbered based on the order of their predictions sorted with respect to the start time (\ie, node 0 represents the instance that starts first). These visualizations demonstrate that the model indeed learns non-linear dependencies between the action instances in a video. The legend below each figure shows the action labels corresponding to the color coded elements. For details on the visualization process, please refer to Section~\ref{subsec:vis_graphs}}.
    \label{fig:vis_graphs}
\end{figure*}
We visualize the learned action query graph in Figure~\ref{fig:vis_graphs}. by observing the graph embeddings obtained from the last layer of decoder. For better visibility, we do not plot the nodes (or their edges) that are classified as no action (\ie class label $\varnothing$) by the prediction head. Note that the edge matrix is also learnable in our model. For the purpose of this visualization, we obtained the edge weights from the attention coefficients in the self-attention based graph message passing module . We show samples with reoccurring and/or overlapping action instances. The visualizations demonstrate that the model indeed learns non-linear dependencies among the action instances that appear in the video.

\subsubsection{Analysis: Effect of Action Instance Durations}\label{subsec:vis_analysis}
\begin{figure}
    \centering
    \includegraphics[width=0.5\textwidth]{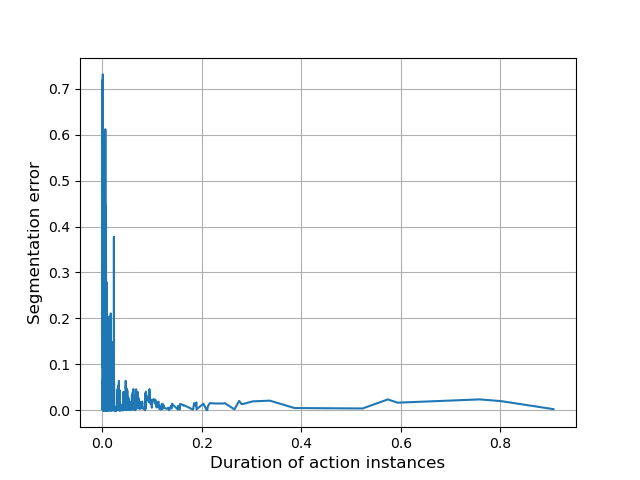}
    \caption{\textbf{Analysis (THUMOS14). }Analysis of segmentation error (L1 loss) with respect to the duration of corresponding ground truth instances. All the values are normalized with respect to the overall video duration. We observe that the action instances of longer durations have lower segmentation errors in their predictions.}
    \label{fig:segmentation}
\end{figure}

We conduct further analysis to study the performance of our model in terms of the durations of the action instances.
Figure~\ref{fig:segmentation} shows the trend of segmentation error, \ie, $L_1$ norm computed between the ground truth and predicted timestamps of actions instances plotted against the duration of the ground truth instances (normalized with respect to the video duration). The error is computed over normalized values of the timestamps. This analysis indicates that action instances with larger durations (with respect to the whole video duration) have lower segmentation errors in their predictions as compared to the instances with smaller durations.

\begin{figure*}
\centering
\includegraphics[width=\textwidth]{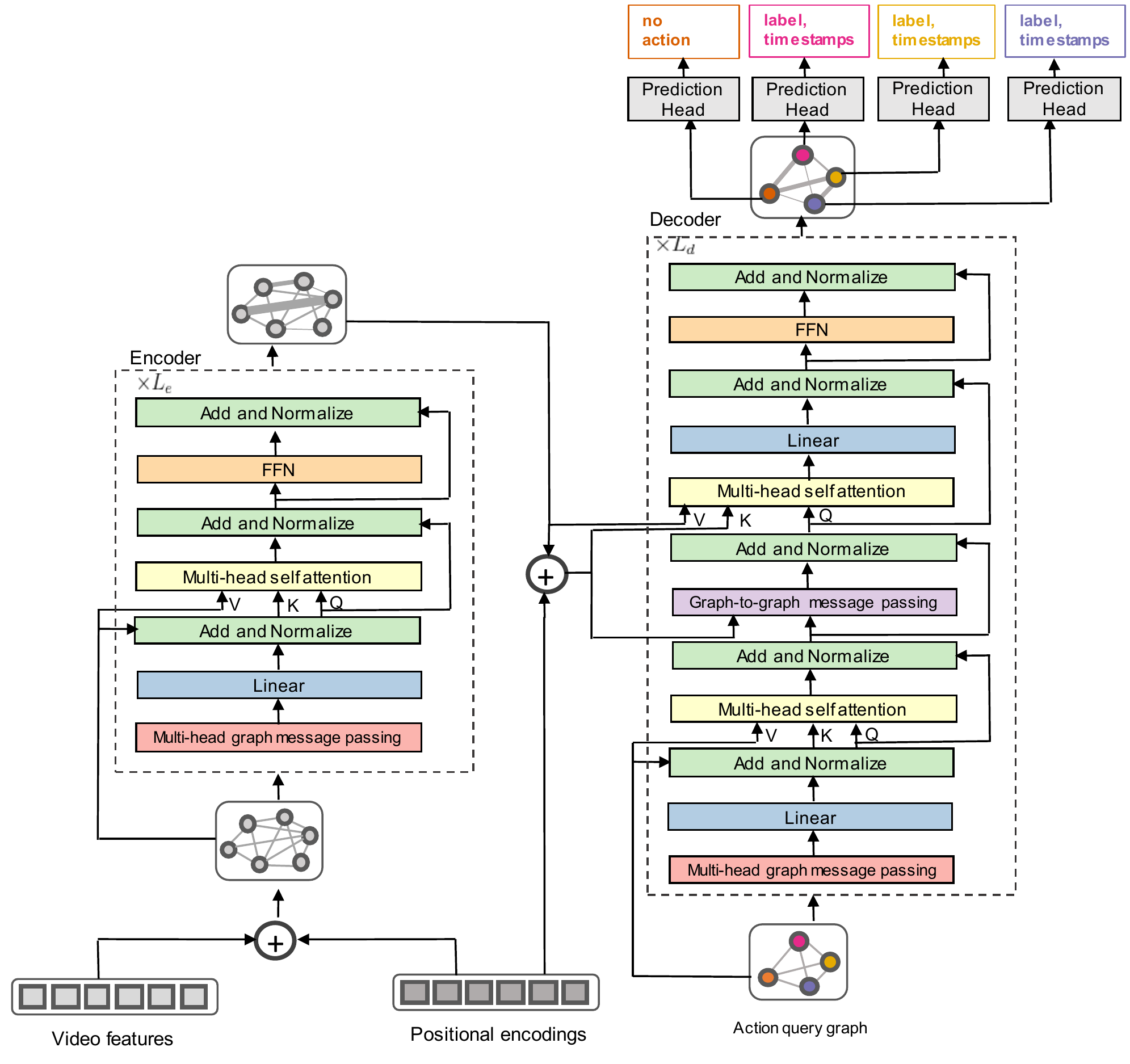}
\caption{\textbf{Detailed Architecture} Architecture of Activity Graph Transformer. Please see Section~\ref{subsec:imp_arch} for details. `Q',`K',`V' are query, key and value to the self-attention layer as described in ~\cite{vaswani2017attention}.}
\label{fig:agt_detail}
\end{figure*}

\subsection{Technical Details}
In this section, we provide additional implementation details to supplement the model section in the main paper.

\subsubsection{Additional details}
\label{subsec:imp_arch}
\textbf{Detailed Architecture.} Figure~\ref{fig:agt_detail} presents the architecture of our AGT in detail. Activity Graph Transformer (AGT) consists of three components: (1) backbone network to obtain features corresponding to the input video; (2) transformer network consisting of an encoder network and a decoder network that operates over graphs; and (3) prediction heads for the final prediction of action instances of the form (label, start time, end time). The encoder network receives the compact video-level representation from the backbone network and encodes it to a latent graph representation, referred to as \textit{context graph}. The decoder network receives graph-structured abstract query encodings (referred to as \textit{action query graph}) as input along with the context graph. The decoder uses the context graph to transform the action query graph to a graph-structured set of embeddings. Each node embedding of this decoder output is fed into a prediction head to obtain predictions of action instances. The whole AGT network is trained end-to-end using a combination of classification and regression losses for the action labels and timestamps respectively. 

\vspace{0.05in}
\noindent
\textbf{Positional Encoding. }Positional encoding layer consists of a layer that retrieves encodings based on an integer index provided to it. In our case, given a video feature $\mathbf{v} = [\mathbf{v}^{(1)},\mathbf{v}^{(2)} \ldots \mathbf{v}^{(N_v)}]$, the positional encoding layer receives input $i$ and provides an embedding $\mathbf{p}_v^{(i)}$ corresponding to the $i$-th element of the video feature $\mathbf{v}^{(i)}$ where $i \in {1,2,\ldots,N_v}$. In our implementation, the embedding size is same as that of the video feature so as to allow addition of the positional encodings and input video features. Since the weights of the layer are learnable during training, the positional encoding layer is learnable. We use \texttt{torch.nn.Embedding} in Pytorch to implement it. This layer initialization requires maximum possible value of $N_v$ in the features corresponding to the video.

\vspace{0.05in}
\noindent
\textbf{Action Query Graph. }Similar to positional encoding layer, the $N_o$ encodings in the action query graph $\mathbf{q}$ is obtained using an embedding layer. Specifically, the layer receives $i$ as input to provide $i$-th node $\mathbf{q}^{(i)}$ of the query graph where $i \in {1,2,\ldots,N_o}$. In our implementation, we use \texttt{torch.nn.Embedding} in Pytorch to implement this. The weights of this layer are learnable during training.

\vspace{0.05in}
\noindent
\textbf{Losses. }For completeness, we describe the IoU loss ($\mathcal{L}_{iou}$) which is used as a component of segment loss $\mathcal{L}_s$ to train our model. The segment loss is described as:
\begin{equation}
    \mathcal{L}_s = \lambda_{iou} \mathcal{L}_{iou}(s^{(i)}, \tilde s^{(\phi(i))})
     + \lambda_{L1} ||s^{(i)} - \tilde s^{(\phi(i))}||_{1} ,
\end{equation}
where $\lambda_{iou}, \lambda_{L1} \in \mathbbm{R}$ are hyperparameters.

\begin{equation}
     \mathcal{L}_{iou}(s^{(i)}, \tilde s^{(\phi(i))}) = 1 - \frac{|s^{(i)} \cap \tilde s^{(\phi(i))}|}{|s^{(i)} \cup \tilde s^{(\phi(i))}|}
\end{equation}
 where $|.|$ is the duration of the instance, \ie, difference between end and start timestamp.

\subsubsection{Training Details}
\label{subsec:imp_tech}
\textbf{Feature Extraction \& Data augmentation.} To obtain I3D features corresponding to an input video $V$ containing $T$ frames sampled at a specific sample rate, we first divide the video into short overlapping segments of 8 frames with an overlap of 4 frames resulting in $T'$ chunks. We use an I3D model pretrained on the Kinetics~\cite{carreira2017quo} dataset to extract features of dimension $C$ ($=2048$). In our implementation, we obtain two-stream features (both RGB and flow streams). We obtain features for these $T'$ chunks to obtain a tensor of size $T' \times 2048$. Here, the length of the video $T$ depends on the duration of the video, and, hence the size of the temporal channel (\ie $T'$) of the feature tensor varies based on the input.

To prevent severe overfitting, we perform data augmentation to train our model on the features directly obtained from I3D model (described above). We use a hyperparameter $N_{v}^{max}$ as the maximum size of temporal channel used for training. This helps in stabilizing the training as the video datasets contain high variance in their duration. If the size of temporal channel of the video tensor $T'$ is less than $N_{v}^{max}$, we repeat each element in the temporal channel $\gamma$ times ($\gamma=4$) in our implementation to obtain a modified tensor of size $\gamma T' \times 2048$ and then randomly sample $T'$ elements from the modified tensor. If the size of temporal channel of the video tensor $T'$ is more than $N_{v}^{max}$, we just randomly sample $T'$ elements from the modified tensor. Note that, positional encoding is applied on this feature of size $N_v = \min(T',N_{v}^{max})$.

 We find such data augmentation during training to be crucial to prevent overfitting and obtain good performance of our model, especially for smaller datasets such as THUMOS14. 
During testing, if the size of temporal channel of the video tensor $T'$ is less than $N_{v}^{max}$, we don't perform any augmentation. If the size of temporal channel of the video tensor $T'$ is more than $N_{v}^{max}$, we uniformly sample $T'$ elements from the feature in order to match the maximum index of the positional encoding layer.

Furthermore, to perform training in minibatches, we apply $0$-padding to ensure all elements have the same size as the largest element of the batch. For training efficiency and minimizing the amount of $0$-padding, we sort all the dataset 
based on the duration of the video. We observe that this type of batch formation leads to improvement in training speed without affecting the model performance.

\vspace{0.05in}
\noindent
\textbf{Hyperparameters.} We provide the hyperparameters used to train our model below.

We train all our models using AdamW optimizer ~\cite{loshchilov2017decoupled} with a learning rate of 1e-5 and a weight decay of 1e-5 for 3000k steps. We reduce the learning rate by factor of 10 after 2000k steps. The hyperparameters in the loss functions $\lambda_{L1}$ and $\lambda_{iou}$ are set to 5 and 3 respectively for all our experiments. All the learnable weights are initialized using Xavier initialization. 

For our experiments, we sample frames at 1/4 of the original frame rate and obtain the I3D features as decribed earlier. We do not finetune the I3D model.

We mention the dataset specific hyperparameters below.

\noindent \textbf{THUMOS14. } We do not use dropout for this dataset. We use maximum number of nodes in the context graph $N_{v}^{max}$ equal to 256. The size of the action query graph is 300 for our experiments (except when conducting ablation on the size of action query graph). We use base model dimension in the transformer as 512 and set the number of encoder and decoder layers as 4 (except when conducting ablation on the number of layers).

\textbf{Charades. } We use dropout with default probability $0.1$. We use maximum number of nodes in the context graph $N_{v}^{max}$ equal to 64. The size of the action query graph is 100 for our experiments (except when conducting ablation on the size of action query graph). We use base model dimension in the transformer as 512 and set the number of encoder and decoder layers as 4 (except when conducting ablation on the number of layers).

\textbf{Epic-Kitchens100. }We do not use dropout for this dataset. We use maximum number of nodes in the context graph $N_{v}^{max}$ equal to 1024. The size of the action query graph is 1200 for our experiments (except when conducting ablation on the size of action query graph). We use base model dimension in the transformer as 512 and set the number of encoder and decoder layers as 4 (except when conducting ablation on the number of layers).



\end{document}